\pdfoutput=1

\documentclass[11pt]{article}

\usepackage[final]{acl}

\usepackage{times}
\usepackage{latexsym}

\usepackage[T1]{fontenc}

\usepackage[utf8]{inputenc}

\usepackage{microtype}

\usepackage{inconsolata}

\usepackage{graphicx}

\usepackage{courier}
\usepackage{amsmath}
\usepackage{amssymb}
\usepackage{bbm}
\usepackage{adjustbox}
\usepackage{pifont} 
\usepackage{xcolor} 
\usepackage{booktabs}
\usepackage{multirow}
\usepackage{multicol}
\usepackage{changepage,threeparttable} 
\usepackage{makecell}
\usepackage{colortbl}
\usepackage{xspace}
\usepackage{enumitem}
\usepackage{subfigure}
\usepackage{wrapfig}
\usepackage{utfsym}

\title{MinosEval: Distinguishing Factoid and Non-Factoid for Tailored Open-Ended QA Evaluation with LLMs}

\author{Yongqi Fan\thanks{~~Equal contribution.}, 
 Yating Wang\footnotemark[1], 
 Guandong Wang \\
 \textbf{Jie Zhai},
 \textbf{Jingping Liu},
 \textbf{Qi Ye}\thanks{~~Corresponding authors.},
 \textbf{Tong Ruan}\footnotemark[2] \\
School of Information Science and Engineering, East China University \\ of Science and Technology, Shanghai, China \\
\texttt{johnnyfans@mail.ecust.edu.cn}, \texttt{ruantong@ecust.edu.cn} \\
}

\begin{document}
\maketitle
\begin{abstract}
Open-ended question answering (QA) is a key task for evaluating the capabilities of large language models (LLMs). Compared to closed-ended QA, it demands longer answer statements, more nuanced reasoning processes, and diverse expressions, making refined and interpretable automatic evaluation both crucial and challenging. Traditional metrics like ROUGE and BERTScore struggle to capture semantic similarities due to different patterns between model responses and reference answers. Current LLM-based evaluation approaches, such as pairwise or listwise comparisons of candidate answers, lack intuitive interpretability. While pointwise scoring of each response provides some descriptions, it fails to adapt across different question contents. Most notably, existing methods overlook the distinction between factoid and non-factoid questions. To address these challenges, we propose \textbf{MinosEval}, a novel evaluation method that first distinguishes open-ended questions and then ranks candidate answers using different evaluation strategies. For factoid questions, it applies an adaptive key-point scoring strategy, while for non-factoid questions, it uses an instance-aware listwise ranking strategy. Experiments on multiple open-ended QA datasets, including self-built ones with more candidate responses to complement community resources, show that MinosEval better aligns with human annotations and offers more interpretable results.
\end{abstract}
\begin{figure*}[ht]
    \centering
    \includegraphics[width=0.88\linewidth]{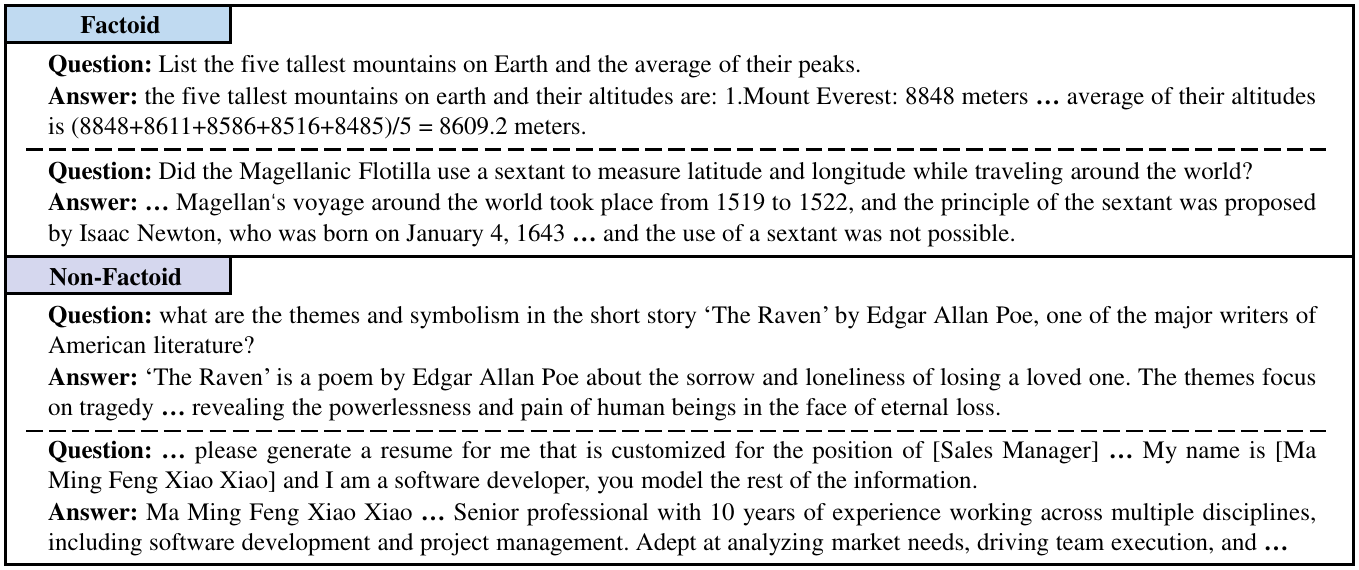}
    \caption{Typical Samples of Factoid and Non-Factoid Open-Ended QA.}
    \label{fig:typicalQA}
\end{figure*}
\section{Introduction}
\label{sec:intro}
Open-ended question answering (QA) is a fundamental task type in comprehensive large language modeling (LLM) evaluation benchmarks and plateforms~\cite{srivastava2023beyond, 2023opencompass, liu2024medbench, myrzakhan2024open, liu2024openeval}. Unlike closed-ended questions, which generally elicit brief, predefined responses (e.g., multiple-choice question~\cite{wang2024mmlu}, true/false question~\cite{luo2023systematic}, and close question), open-ended questions~\cite{kantharaj2022opencqa, tao2024chain, lin2024asmr} necessitate the generation of longer, more detailed answers that require complex reasoning, nuanced understanding, and diverse modes of expression. Consequently, the evaluation of open-ended QA has become a significant area of research, with an increasing focus on developing methods and benchmarks~\cite{amirizaniani2024can, yang2024linkage, wang-etal-2024-revisiting} that not only measure the quality of the model responses but also provide interpretability and alignment with human annotation.

Traditional evaluation metrics for free-text responses, such as BLEU~\cite{papineni2002bleu}, ROUGE~\cite{lin2004rouge}, and BERTScore~\cite{zhangbertscore}, focus on lexical overlap or semantic similarity. While useful for tasks with clear answers, they are less effective for open-ended QA, where responses vary in style and content. These metrics also struggle with complex open-ended answers, involving reasoning, creativity, and entailment.

In recent years, evaluation methods have shifted toward leveraging LLMs for automatic evaluation. These methods typically involve employing pairwise comparisons for Elo~\cite{boubdir2023elo, harang2024beyond}, listwise comparisons where LLMs rank responses based on relevance or quality, and pointwise approaches~\cite{liangholistic, vu2024foundational} that score responses on predefined dimensions. Meanwhile, some studies conducted pairwise and pointwise supervised fine-tuning to obtain dedicated LLMs for evaluation~\cite{kim2024prometheus, li2023generative}. While promising, these methods have notable limitations. Pairwise methods involve multiple comparisons, leading to $O(n^2)$ complexity for $n$ responses. Pointwise methods, though scoring on predefined criteria, e.g., ``Fluency'' and ``Truthfulness'', often fail to be adapted to each specific question context. Additionally, pairwise and pointwise methods suffer from the tie-breaking problem. In addition, pairwise and listwise methods lack intuitive explanations, making their rankings difficult for human evaluators to interpret.

More notably, these LLM-based approaches overlook the distinction between factoid and non-factoid questions in open-ended QA. Factoid~\cite{jiang2019freebaseqa, gaikwad2023factoid} and non-factoid~\cite{lakshmi2023study, bolotova-baranova-etal-2023-wikihowqa} questions exhibit a clear distinction. Factoid questions aim to elicit answers based on objective, real-world facts or entities, with a relatively fixed and clear scope, e.g., ``\textit{List the five highest mountains on Earth and the average elevation of their peaks}''. Of course, additional sentences can be included in the answer to ensure semantic coherence. In contrast, non-factoid questions offer more creative freedom, with the primary requirement being to meet the demands of the question without necessarily including critical, fact-based information, e.g., ``\textit{Please help me write homophone joke}''. By failing to account for this distinction, such a one-size-fits-all strategy undermines the precision and effectiveness of evaluation.

In this paper, we propose \textbf{MinosEval}, a novel two-stage approach for evaluating open-ended QA and providing corresponding interpretive information. It distinguishes between factoid and non-factoid questions based on their semantic and content differences, employing tailored evaluation strategies for each type of question. For factoid questions, we use an adaptive key-point scoring strategy that extracts key factoid points from a given reference answer. It then compares how each response entails these key points using a natural language inference (NLI) model, inspired by~\citet{bohnet2022attributed}. For non-factoid questions, we apply an instance-aware listwise ranking approach, generating five levels of silver answer instances to enhance the LLM's performance in directly ranking these more creative and diverse responses.
\begin{table*}[ht]
\centering
\small
\caption{Dataset statistics and description. The columns indicate the dataset name, source, language, number of reference answers, samples, and candidate model responses to be evaluated.
}
\label{tab:datasets}
\resizebox{\textwidth}{!}{
    \begin{tabular}{l|ccccccc}
        \toprule
        \multirow{2}{*}{Datasets} & \multirow{2}{*}{Source} & \multirow{2}{*}{Language} & \multirow{2}{*}{\# Res} & \multirow{2}{*}{\# Ref} & \multicolumn{3}{c}{\# Samples} \\ 
        \cmidrule(lr){6-8}
        & & & & & Factoid & Non-factoid & Total \\ 
        \midrule
        ANTIQUE\_S5      & ANTIQUE~\cite{hashemi2020antique}  & English & 2\text{ to }4 & 2 & 96$\times$5  & 404$\times$5 & 500$\times$5 \\
        TREC-DL-NF\_S5   & TREC-DL-NF~\cite{craswell2020overview, craswell2021overviewtrec2020deep}  & English & 2\text{ to }4  & 1 & 28$\times$5  & 27$\times$5 & 55$\times$5 \\
        AlignBench\_Minos   & AlignBench~\cite{liu2023alignbench} & Chinese & 6  & 1   & 299 & 384 & 683 \\
        GaokaoBench\_Minos  & GaokaoBench~\cite{zhang2023evaluating} & Chinese & 6 & 1 & 158 & 247 & 405 \\
        \bottomrule
    \end{tabular}
}
\end{table*}

Our approach tackles the challenges of existing methods in open-ended QA evaluation and offers several advantages. First, it distinguishes between different types of open-ended questions, enabling a more tailored evaluation strategy. It also provides clear ranking guidelines based on the characteristics of factoid and non-factoid questions, including key points and silver instances to enhance interpretability. Moreover, the guidelines are adaptive to the specific questions, and the entire process is fully automated, requiring no manual intervention.

We conducted extensive experiments on four datasets. Following the approach of~\citet{yang2024linkage}, we performed five sets of sampling to construct ANTIQUE\_S5 and TREC-DL-NF\_S5 from the publicly available open quality assurance datasets ANTIQUE~\cite{hashemi2020antique} and TREC-DL-NF~\cite{craswell2020overview, craswell2021overviewtrec2020deep}. Additionally, we created two self-built datasets, AlignBench\_Minos and GaokaoBench\_Minos, which contain a larger number of candidate responses based on AlignBench~\cite{liu2023alignbench} and GaokaoBench~\cite{zhang2023evaluating} to supplement the resources of the research community. The results demonstrate that our method outperforms existing LLM-based evaluation approaches for open-ended QA. We also present case studies in which the key points for factoid questions and the silver answer instances for non-factoid questions provide a valuable interpretive foundation.

We hope this work contributes to the effective evaluation of LLMs' performance in open-ended QA and promotes further research within the LLM community. Our datasets, evaluation results, and code for \textbf{MinosEval} are publicly available\footnote{\url{https://github.com/JOHNNY-fans/MinosEval}}.
\section{Related Work}
\subsection{Open-ended QA.}
Question answering (QA) is a key task in natural language processing, aimed at providing accurate answers to satisfy the user's information need or request. Depending on the scope of the answer, QA systems are typically divided into closed-ended and open-ended types. Closed-ended QA includes formats like multiple choice~\cite{wang2024mmlu}, true/false~\cite{luo2023systematic}, and close question~\cite{yu2023alert}, while open-ended QA allows for more diverse, unbounded responses, typically divided into factoid and non-factoid questions~\cite{agustianingsih2019design}. Common tasks in open-ended QA include reading comprehension~\cite{liu2023webglm}, summarization~\cite{fabbri2021summeval}, and writing~\cite{ngo2024effectiveness}.

The advent of large language models (LLMs) such as ChatGPT~\cite{Chatgpt} and Gemini~\cite{gemini} has made open-ended QA a popular task in LLM evaluation benchmarks. This task is essential for evaluating the ability of LLMs to generate complex, creative, and contextually relevant responses, which require advanced reasoning. Moreover, open-ended QA has been used to explore LLM limitations in understanding nuanced language~\cite{dentella2024testing} and capturing human intent and emotions~\cite{amirizaniani2024can}, further highlighting its importance as a benchmark for evaluating reasoning and cognitive abilities~\cite{yang2024linkage, wang2024revisiting}.
\subsection{Open-ended QA Evaluation}
\textbf{Traditional Methods:}
Matching-based evaluation methods for open-ended QA include BLEU~\cite{papineni2002bleu} and ROUGE~\cite{lin2004rouge}. These methods primarily focus on matching the n-grams between the generated and reference texts, which often overlooks the semantic meaning. For example,~\cite{krishna-etal-2021-hurdles} highlights that ROUGE is ineffective for long-text QA tasks.
LM-based methods aim to capture semantic similarity better. BERTScore~\cite{zhangbertscore} easuring semantic similarity between embeddings. Further work introduces BEM~\cite{bulian2022tomayto}, a BERT-based model for assessing semantic equivalence between candidate and reference answers, and PEDANTS~\cite{li2024pedants}, which employs rule-guided rubrics and lightweight neural scoring. However, they struggle with complex questions and diverse answers.

\textbf{Human Evaluation:}
Human evaluation remains the golden standard in open-ended QA~\cite{bolotova2023wikihowqa}, providing more accurate and comprehensive feedback~\cite{chang2024survey}. However, it is resource-intensive and difficult to scale, limiting its use in large-scale evaluations.
\begin{figure*}[ht]
    \centering
    \includegraphics[width=0.88\linewidth]{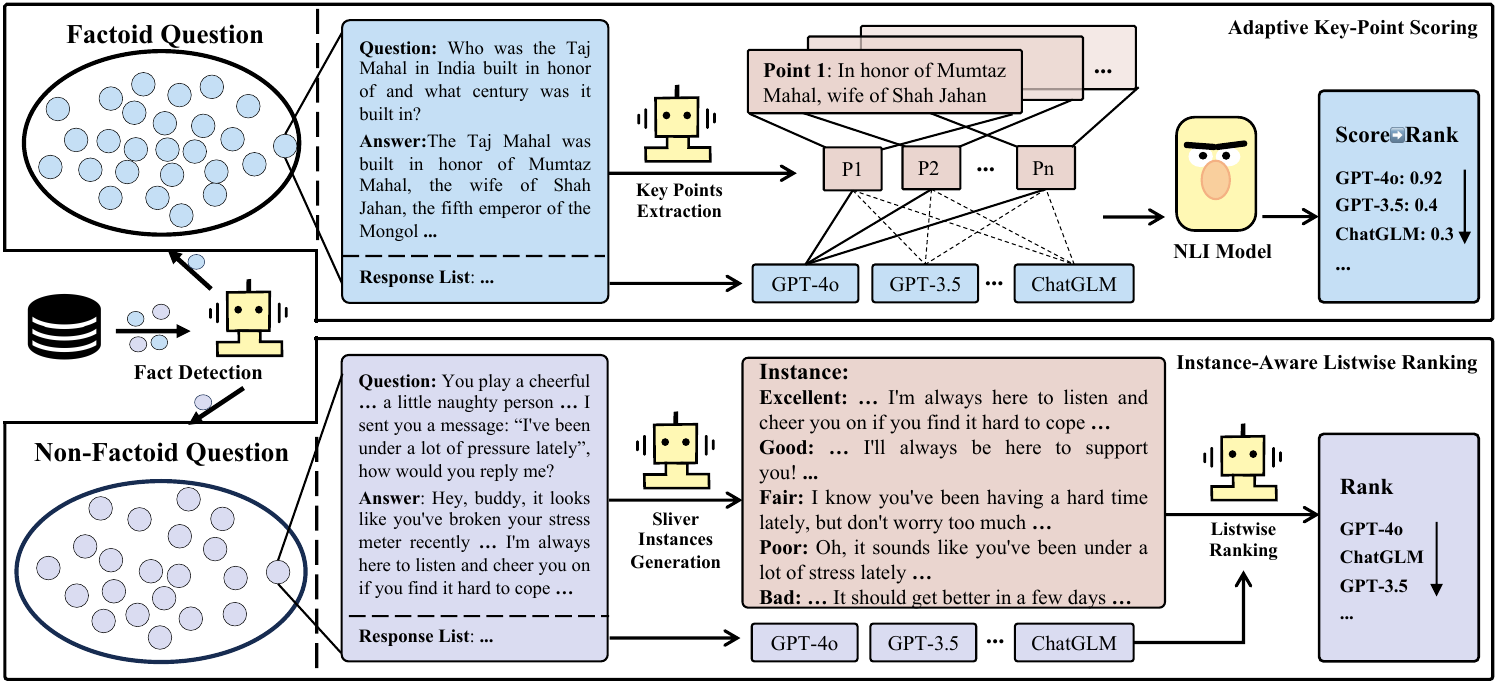}
    \caption{The MinosEval Workflow: Distinguishing Factoid vs. Non-Factoid Questions, Adaptive Key-Point Scoring, and Instance-Aware Listwise Ranking.}
    \label{fig:framework}
\end{figure*}

\textbf{LLM-based Evaluation:}
With the development of LLMs, they have demonstrated evaluation capabilities similar to human annotations~\cite{li2023collaborative}. Evaluation methods based on LLMs primarily include pointwise~\cite{liu2023g}, pairwise~\cite{shi2024judging}, and listwise approaches. Recent research has explored these methods, such as using textual entailment to evaluate model responses in open-ended QA~\cite{yao2024accurate}. PERSE~\cite{wang2024learningpersonalizedalignmentevaluating} combines pointwise and pairwise methods for story generation, while MATEval~\cite{li2024mateval} simulates human collaborative discussions and integrates multiple agents to evaluate open-ended text. While LLM-based evaluation offers flexibility, it faces challenges, including positional bias~\cite{shi2024judging}, verbosity bias, self-bias~\cite{wang2023large}, and efficiency issues. 

Meanwhile, several studies have noticed the distinction between factoid and non-factoid questions. For factoid questions, some work~\cite{min2023factscore, cook2024ticking, yan2024atomic, hu2024refchecker} has attempted to decompose the answers into discrete facts or knowledge triplets, and then build a validation checklist. In contrast, for non-factoid questions, there are some expert judges~\cite{kim2024prometheus, li2023generative} with specialized training to evaluate LLM performance have been proposed.
\section{Methodology}
In this section, we begin by formally defining the open-ended QA evaluation task, explaining the principles of distinguishing between factoid and non-factoid questions, and then presenting the details of \textbf{MinosEval}.
\subsection{Preliminary}
\textbf{Task Definition.} 
Given an open-ended question \( q \) and the corresponding \( n \) candidate model responses \( \mathbb{R} = \{r_1, r_2, \dots, r_n\} \), where \( r_i \) represents the \( i \)-th model response to be evaluated, the goal is to implement a specific strategy and produce a ranked order of these model responses. The reference answer for question \( q \) is denoted as \( a \). Furthermore, it is important to note that the final ranked order strictly avoids ties in this paper.

\noindent \textbf{Factoid vs. Non-Factoid Questions.} 
\label{sec: factoid}
Distinguishing between factoid and non-factoid open-ended questions is a central motivation of this paper. To this end, we provide clear definitions, which will also guide the human annotation of the datasets. Building on previous works~\citet{jiang2019freebaseqa} and~\citet{hashemi2020antique}, we define the following: First, in open-ended QA, the answer is arbitrary free text, not restricted to predefined items or fixed responses. Second, for factoid questions, the answer is expected to contain explicit information, such as entities, common knowledge, and facts. Alternatively, due to the limitations of the question, the scope of the key elements of the answer is narrowly defined. In contrast, non-factoid questions do not require key factoid information and are less constrained in terms of answer content. It is worth noting that in our setting, the additional text included in the model responses, e.g., reasoning and thought processes, does not affect the factoid nature of the question. We present some typical samples of factoid and non-factoid QA in Figure~\ref{fig:typicalQA} to facilitate a more intuitive understanding.
\begin{table*}[ht]
\small
\centering
\caption{Performance on AlignBench\_Minos and GaokaoBench\_Minos: Kendall's Tau (K), Spearman's Rho (S), and Rank-biased Overlap (RBO) for $p$=0.5 and $p$=0.9. ``$\dagger$'' denotes manually classifying factoid and non-factoid questions.}
\label{tab:zh_main_exp}
\resizebox{0.72\textwidth}{!}{
  \begin{tabular}{clcccccccc}
    \toprule
    & \multirow{2}{*}{Method} & \multicolumn{4}{c}{AlignBench\_Minos} & \multicolumn{4}{c}{GaokaoBench\_Minos} \\
    \cmidrule(lr){3-6} \cmidrule(lr){7-10}
    & & K & S & \makecell[c]{RBO \\ ($p$=0.5)} & \makecell[c]{RBO \\ ($p$=0.9)} & K & S & \makecell[c]{RBO \\ ($p$=0.5)} & \makecell[c]{RBO \\ ($p$=0.9)} \\
    \midrule
    \multirow{2}{*}{\makecell[c]{\text{Automatic} \\ \text{Metrics}}} 
    & BLEU & 15.41 & 19.47 & 36.24 & 79.98 & 11.08 & 12.85 & 36.78 & 79.81 \\
    & ROUGE-L & 13.38 & 16.69 & 37.93 & 80.09 & 4.89 & 5.81 & 35.15 & 78.95 \\\midrule
    \multirow{1}{*}{\makecell[c]{\text{LM-based Metrics}}}  
    & BERTScore & 13.62 & 18.12 & 37.81 & 80.15 & 12.92 & 16.44 & 40.21 & 80.66 \\
    \midrule
    \multirow{3}{*}{\makecell[c]{\text{Naive} \\ \text{LLM Evaluation}}} 
    & Pointwise & 32.94 & 41.37 & 47.41 & 83.59 & 31.06 & 38.55 & 42.64 & 82.43 \\
    & Pairwise & 38.66 & 47.89 & 51.59 & 84.85 & 45.71 & 54.93 & 56.90 & 86.45 \\
    & Listwise & 41.47 & 51.46 & 54.53 & 85.64 & 55.33 & 64.66 & 61.59 & 88.35 \\
    \midrule
    \multirow{1}{*}{\makecell[c]{\text{Others}}} 
    & LINKAGE & 35.75  & 43.82 & 52.97 & 84.88  & 37.45 & 44.61 & 52.47 & 84.86 \\
    \midrule
    \multirow{2}{*}{\makecell[c]{\text{Ours}}}
    & MinosEval & 45.28  & 54.89  & 56.30 & 86.28  & 56.12 & 65.77 & 63.36 & 88.67 \\
    & \cellcolor{blue!8}MinosEval$^\dagger$ & \cellcolor{blue!8}\textbf{47.68}  & \cellcolor{blue!8}\textbf{57.38}  & \cellcolor{blue!8}\textbf{57.09} & \cellcolor{blue!8}\textbf{86.62}  & \cellcolor{blue!8}\textbf{59.77} & \cellcolor{blue!8}\textbf{70.30} & \cellcolor{blue!8}\textbf{65.26} & \cellcolor{blue!8}\textbf{89.31} \\
    \bottomrule
  \end{tabular}
}
\end{table*}

\begin{table*}[ht]
\centering
\small
\caption{Performance on ANTIQUE\_S5 and TREC-DL-NF\_S5: Kendall's Tau (K), Spearman’s Rho (S), and Rank-biased Overlap (RBO) for $p$=0.5 and $p$=0.9. ``$\dagger$'' denotes manually classifying factoid and non-factoid questions.}
\label{tab:en_main_exp}
\resizebox{\textwidth}{!}{
  \begin{tabular}{clcccccccc}
    \toprule
    & \multirow{2}{*}{Method} & \multicolumn{4}{c}{ANTIQUE\_S5} & \multicolumn{4}{c}{TREC-DL-NF\_S5} \\
    \cmidrule(lr){3-6} \cmidrule(lr){7-10}
    & & K & S & \makecell[c]{RBO \\ ($p$=0.5)} & \makecell[c]{RBO \\ ($p$=0.9)} & K & S & \makecell[c]{RBO \\ ($p$=0.5)} & \makecell[c]{RBO \\ ($p$=0.9)} \\
    \midrule
    \multirow{2}{*}{\makecell[c]{\text{Automatic} \\ \text{Metrics}}} 
    & BLEU & 26.08\textsubscript{$\pm$ 0.0247} & 29.02\textsubscript{$\pm$ 0.0249} & 66.15\textsubscript{$\pm$ 0.0086} & 91.85\textsubscript{$\pm$ 0.0023} & 28.11\textsubscript{$\pm$ 0.0266} & 31.40\textsubscript{$\pm$ 0.0356} & 65.42\textsubscript{$\pm$ 0.0218} &  91.39\textsubscript{$\pm$ 0.0046} \\
    & ROUGE-L & 22.51\textsubscript{$\pm$ 0.0330} & 25.45\textsubscript{$\pm$ 0.0325} & 64.94\textsubscript{$\pm$ 0.0111} & 91.59\textsubscript{$\pm$ 0.0027} & 34.56\textsubscript{$\pm$ 0.0535} & 37.61\textsubscript{$\pm$ 0.0616} & 68.30\textsubscript{$\pm$ 0.0351} & 92.07\textsubscript{$\pm$ 0.0081} \\
    \midrule
    \multirow{3}{*}{\makecell[c]{\text{LM-based} \\ \text{Metrics}}} 
    & BERTScore & 29.56\textsubscript{$\pm$ 0.0165} & 33.10\textsubscript{$\pm$ 0.0164} & 66.94\textsubscript{$\pm$ 0.0132} & 92.02\textsubscript{$\pm$ 0.0028} & 44.62\textsubscript{$\pm$ 0.0672} & 50.23\textsubscript{$\pm$ 0.0723} & 71.61\textsubscript{$\pm$ 0.0382} &  93.01\textsubscript{$\pm$ 0.0088} \\
    & BEM & 39.75\textsubscript{$\pm$ 0.0178} & 43.37\textsubscript{$\pm$ 0.0221} & 76.32\textsubscript{$\pm$ 0.0073}  & 93.62\textsubscript{$\pm$ 0.0019} & 46.42\textsubscript{$\pm$ 0.0528} & 51.93\textsubscript{$\pm$ 0.0491} & 75.05\textsubscript{$\pm$ 0.0145} & 93.21\textsubscript{$\pm$ 0.0033} \\
    & PEDANTS & 37.85\textsubscript{$\pm$ 0.0206} & 41.49\textsubscript{$\pm$ 0.0218} & 74.03\textsubscript{$\pm$ 0.0070} & 93.18\textsubscript{$\pm$ 0.0016} & 48.12\textsubscript{$\pm$ 0.0505} & 53.82\textsubscript{$\pm$ 0.0393} & 74.61\textsubscript{$\pm$ 0.0189} & 93.31\textsubscript{$\pm$ 0.0044}  \\
    \midrule
    \multirow{3}{*}{\makecell[c]{\text{Naive LLM} \\ \text{Evaluation}}} 
    & Pointwise & 50.36\textsubscript{$\pm$ 0.0155} & 55.99\textsubscript{$\pm$ 0.0150} & 77.08\textsubscript{$\pm$ 0.0053} & 94.42\textsubscript{$\pm$ 0.0012} & 54.79\textsubscript{$\pm$ 0.0335} & 60.33\textsubscript{$\pm$ 0.0282} & 73.58\textsubscript{$\pm$ 0.0171} & 93.37\textsubscript{$\pm$ 0.0051} \\
    & Pairwise & 63.53\textsubscript{$\pm$ 0.0113} & 68.91\textsubscript{$\pm$ 0.0093} & 81.45\textsubscript{$\pm$ 0.0037} & 95.46\textsubscript{$\pm$ 0.0003} & 63.21\textsubscript{$\pm$ 0.0269} & 69.93\textsubscript{$\pm$ 0.0269} & 77.50\textsubscript{$\pm$ 0.0175} & 94.40\textsubscript{$\pm$ 0.0045} \\
    & Listwise & 62.55\textsubscript{$\pm$ 0.0359} & 68.56\textsubscript{$\pm$ 0.0317} & 83.30\textsubscript{$\pm$ 0.0168} & 95.93\textsubscript{$\pm$ 0.0041} & 65.82\textsubscript{$\pm$ 0.0463} & 72.95\textsubscript{$\pm$ 0.0344} & 79.47\textsubscript{$\pm$ 0.0403} & 94.93\textsubscript{$\pm$ 0.0085} \\
    \midrule
    Others & LINKAGE & 52.65\textsubscript{$\pm$ 0.0201} & 57.71\textsubscript{$\pm$ 0.0169} & 79.89\textsubscript{$\pm$ 0.0108} & 95.17\textsubscript{$\pm$ 0.0027} & 65.58\textsubscript{$\pm$ 0.0335} & 72.29\textsubscript{$\pm$ 0.0238} & 79.36\textsubscript{$\pm$ 0.0333} & 94.89\textsubscript{$\pm$ 0.0066} \\
    \midrule
    \multirow{2}{*}{\makecell[c]{\text{Our}}} & MinosEval & 64.93\textsubscript{$\pm$ 0.0075} & 68.83\textsubscript{$\pm$ 0.0061} & 84.69\textsubscript{$\pm$ 0.0064} & 96.19\textsubscript{$\pm$ 0.0015} & 65.45\textsubscript{$\pm$ 0.0213} & 69.56\textsubscript{$\pm$ 0.0227} & 82.03\textsubscript{$\pm$ 0.0210} & 95.31\textsubscript{$\pm$ 0.0041}  \\  
    & \cellcolor{blue!8}MinosEval$^\dagger$ & \cellcolor{blue!8}\textbf{65.97\textsubscript{$\pm$ 0.0097}} & \cellcolor{blue!8}\textbf{69.91\textsubscript{$\pm$ 0.0102}} & \cellcolor{blue!8}\textbf{84.79\textsubscript{$\pm$ 0.0075}}   &
    \cellcolor{blue!8}\textbf{96.27\textsubscript{$\pm$ 0.0016}} & \cellcolor{blue!8}\textbf{68.61\textsubscript{$\pm$ 0.0129}} & \cellcolor{blue!8}\textbf{73.09\textsubscript{$\pm$ 0.0248}} & \cellcolor{blue!8}\textbf{84.38\textsubscript{$\pm$ 0.0164}} & \cellcolor{blue!8}\textbf{95.82\textsubscript{$\pm$ 0.0036}} \\
    \bottomrule
  \end{tabular}
}
\end{table*}
\subsection{MinosEval}
Figure~\ref{fig:framework} shows how our MinosEval works. Specifically, given an open-ended question \( q \), a reference answer \( a \), and model responses \( \mathbb{R} \) that need to be evaluated, this sample is first distinguished by an LLM-based fact detection module into either factoid cluster \( C_f \) or non-factoid cluster \( C_{nf} \). For factoid samples, the final rank orders \( \mathcal{R}_{f} \) are generated by applying the adaptive key-point scoring strategy. For non-factoid samples, the final rank orders \( \mathcal{R}_{nf} \) are produced by the instance-aware listwise ranking strategy. The implementation details of each module are as follows.
\subsubsection{LLM-based Fact Detection}
In light of the differences between factoid and non-factoid open-ended QA discussed in Section~\ref{sec: factoid}, it is necessary to treat these two types of questions differently. However, manual annotation is costly, so instead, we leverage the instruction-following and few-shot learning capabilities of advanced LLMs~\cite{brown2020language}, such as GPT-4o~\cite{GPT-4o}. Specifically, we employ an in-context learning approach to develop a simple yet effective fact detection module, which classifies samples based on the questions and reference answers by designing the task prompt and providing suitable demonstration examples. The specific prompt is shown in Figure~\ref{fig:llmdisting_zh} in the Appendix.
\subsubsection{Adaptive Key-Point Scoring}
For factoid open-ended questions, the reference answer typically contains facts, entities, or common knowledge, or is constrained by the question's content, such as in rewriting tasks. These key points must be considered during ranking, so we propose an adaptive key-point scoring strategy. Key information is extracted from the reference answers, but unlike conventional pointwise evaluation, it adapts to the specific question rather than fixed criteria. We designed an LLM-based key-point extraction module, with the prompt provided in Figure~\ref{fig:keypoints_en} in the Appendix.

Having obtained these key points, we proceed to calculate scores for each model response to be evaluated. This is modeled as a Natural Language Inference (NLI) task, where the degree of entailment and contradiction between each model response and each key point are computed. The final ranking of the model responses is then determined based on these scores. The formal definition is shown in the Formula~\ref{formula:Rf}.
\begin{equation}
\small
    \mathcal{R}_f = \{\text{Sort}\left(S_f(q_i, a_i, \mathbb{R}_i), \mathbb{R}_i\right) \mid q_i, a_i, \mathbb{R}_i \in C_f\}
\label{formula:Rf}
\end{equation}

where \( \mathcal{R}_f \) represents the ranking results of samples in the factoid cluster \( C_f \), \( q_i \) denotes the \( i \)-th question, \( a_i \) denotes the \( i \)-th reference answer, and \( \mathbb{R}_i \) refers to the list of candidate model responses. \( \text{Sort} \) denotes a simple sorting function that sorts \( \mathbb{R}_i \) based on scores, and \( S_f \) is the scoring function that applies the adaptive key-point scoring strategy, which is specifically shown in the Formula~\ref{formula:S_f}.
\begin{equation}
\small
    \begin{aligned}
    S_f(q, a, \mathbb{R}) = \{\frac{1}{|\mathbb{K}|} \sum_{k_j \in \mathbb{K}} \left( \text{NLI}\left (r_i, k_j \right) \right) \mid r_i \in \mathbb{R} \},\\ \text{note } \mathbb{K} = \text{extractKeyPoints}\left(q, a\right)
    \end{aligned}
\label{formula:S_f}
\end{equation}

where \( S_f(q, a, \mathbb{R}) \) is computed by extracting key points \( \mathbb{K} \) from question \( q \) and reference answer \( a \). and then calculating the NLI score between each model response \( r_i \in \mathbb{R} \) and each key point \( k_j \in \mathbb{K} \). The final score for each model response is the average of its NLI scores across all key points. The definition of the NLI function is shown in Formula~\ref{formula:NLI}.
\begin{equation}
\small
    \text{NLI}\left (r_i, k_j \right) = s_{e\_ij} - s_{c\_ij}, \text{where } r_i \in \mathbb{R}, k_j \in \mathbb{K}
\label{formula:NLI}
\end{equation}

where $\text{NLI}(r_i, k_j)$ is the entailment probability of the model response \( r_i \in \mathbb{R} \) with the key point \( k_j \in \mathbb{K} \), minus the contradiction probability. Here, \( s_{e\_ij} \) denotes the entailment probability between the \( i \)-th response \( r_i \) and the \( j \)-th key point \( k_j \), while \( s_{c\_ij} \) represents the contradiction probability.
\subsubsection{Instance-Aware Listwise Ranking}
For non-factoid open-ended questions, answers are not constrained by fixed key information or the content of the question but focus on creative expression based on meaning. As a result, comparing model responses becomes crucial. We adopt the classic listwise approach, introducing an LLM ranker \( A_{nf} \) to rank candidate responses based on the question and reference answer.

To further improve the stability and accuracy of the ranking, we propose an instance-aware listwise ranking strategy, the formal definition is shown in the Formula~\ref{formula:Rnf}. Specifically, we use LLMs to automatically generate silver instances of varying quality levels based on the question and reference answer, using them to enhance the performance of listwise ranking. The ``silver'' means that these instances have not been modified manually. Specific prompts for generating silver instances and LLM-based listwise ranking are shown in Figure \ref{fig:instance} and Figure \ref{fig:list_wise_instance} in the Appendix.
\begin{equation}
\small
    \begin{aligned}
    \mathcal{R}_{nf} =\{A_{nf}\left(q_i, a_i, \mathbb{I}_i, \mathbb{R}_i\right) \mid q_i, a_i, \mathbb{R}_i \in C_{nf}\},\\ \text{note } \mathbb{I}_i = \text{generateInstance}\left(q_i, a_i\right)
    \end{aligned}
\label{formula:Rnf}
\end{equation}

where \( \mathcal{R}_{nf} \) represents the ranking results of the non-factoid cluster \( C_{nf} \), \( q_i \) denotes the \( i \)-th question, \( a_i \) denotes the \( i \)-th reference answer, and \( \mathbb{R}_i \) refers to the \( i \)-th candidate model responses list. \( \mathbb{I}_i \) is the generated sliver instances for the \( i \)-th sample. \( A_{nf} \) is the LLM ranker used for listwise ranking.

\section{Experiments}
\label{sec:ex}
\subsection{Dataset}
We filter and manually annotate the four open datasets to conduct four open-ended QA evaluation datasets. The statistics are shown in Table~\ref{tab:datasets}.

\textbf{AlignBench\_Minos} and \textbf{GaokaoBench\_Minos}: AlignBench~\cite{liu2023alignbench} is an open-ended QA benchmark with real-scenario rooted queries and corresponding human verified references. GaokaoBench~\cite{zhang2023evaluating} uses Chinese National College Entrance Examination (GAOKAO) questions as a dataset, we select a subset of subjective questions. For both datasets, we generated six LLM responses for each sample and annotated their ranked orders to obtain our evaluation datasets. The specific LLMs used are: GPT-4o~\cite{GPT-4o}, GPT-3.5~\cite{GPT-3.5}, LLAMA~\cite{touvron2023llama}, ChatGLM~\cite{glm2024chatglm}, InternLM~\cite{cai2024internlm2}, and Qwen~\cite{yang2024qwen2}.

\textbf{ANTIQUE\_S5} and \textbf{TREC-DL-NF\_S5}: ANTIQUE~\cite{hashemi2020antique} and TREC-DL-NF~\cite{craswell2020overview, craswell2021overviewtrec2020deep} are classic open-ended QA datasets. Following the experimental setup of~\citet{yang2024linkage}, we sampled five subsets of each dataset (denoted as \textbf{S5}) to construct our experimental datasets. Notably, although these datasets were originally classified as non-factoid, we revisited their samples based on the rules outlined in Section~\ref{sec: factoid}. During this process, we identified samples involving common sense and facts, which were subsequently reclassified. We list some reclassified samples in Table~\ref{tab:trec_antique} in the Appendix.

\textbf{Annotation}: We conducted two annotation tasks for the open-ended QA datasets, AlignBench\_Minos and GaokaoBench\_Minos, where we annotated the gold rank orders based on the quality of candidate model responses. The team included one PhD student, two Master's students (specializing in NLP), and an LLM annotator (GPT-4o). Each participant initially performed independent annotations, followed by discussions to reach a consensus. Human annotators had access to external knowledge via the Internet. Our statistics show that in 98.39\% of cases, the rankings produced by GPT-4o were modified by human annotators, emphasizing the need to re-interpret the annotation process for greater reliability. Additionally, we categorized the samples based on the rules for distinguishing factoid from non-factoid open-ended QA, as described in Section~\ref{sec: factoid}.
\subsection{Baselines}
We compare traditional metrics BLEU~\cite{papineni2002bleu}, ROUGE~\cite{lin2004rouge}, and BERTScore~\cite{zhangbertscore}, as well as BEM~\cite{bulian2022tomayto} and PEDANTS~\cite{li2024pedants}. Due to language constraints, PEDANTS and BEM are only compared on English datasets. In addition, we evaluate three naive LLM evaluation methods, including Pointwise, Pairwise, and Listwise, together with the recent LINKAGE approach~\cite{yang2024linkage}. We also compare methods and supervised models dedicated to evaluating factoid and non-factoid questions, including FActScore~\cite{min2023factscore}, RefChecker~\cite{hu2024refchecker}, PROMETHEUS 2~\cite{kim2024prometheus}, and AUTO-J~\cite{li2023generative}. Based on the categorization and applicable languages of these methods, we selected two datasets for our experiments.

For all LLM-based methods in the main experiments, GPT-4 was used as the base model. To ensure comparative fairness, tie-breaking situations were addressed by analyzing the output of each method and processing it to determine the final ranking, as detailed in Section~\ref{app:tie-breaking}. The input, output, and parsing processes are summarized in Table~\ref{tab:input_output} in Appendix~\ref{app:rank_process}. Part of the code is based on an open-source repository\footnote{\url{https://github.com/babyyang525/LINKAGE-Listwise-NFQA-Evaluation}}. The specific prompts of these methods are shown in Figures~\ref{fig:point_wise} to~\ref{fig:refchecker-checker}.
\subsection{Evaluation Metrics}
To evaluate the effectiveness of open-ended QA evaluation, we employ Kendall's Tau, Spearman's Rho, and Rank-Biased Overlap (RBO)~\cite{webber2010similarity} to measure the alignment between model-generated and human-annotated ranked orders. Among these metrics, \textbf{Spearman's Rho} is selected as the primary metric due to its balance between robustness and sensitivity to monotonic relationships. RBO serves as a supplementary metric, with its parameter \( p \) adjusting the weighting of the rank positions. Lower \( p \) values place greater emphasis on higher-ranked items, thereby prioritizing the evaluation of top positions.
\subsection{Implementation Details}
We inferred open-source LLMs and NLI models on a single NVIDIA A100 80GB GPU using the official deployment method, and for closed-source commercial LLMs, we used the official APIs to obtain responses. Meanwhile, we set the inference temperature as 0 and the random seed to 42 to eliminate randomness. The specific version numbers of all LLMs used for the experiments are listed in Table~\ref{tab:llms_version} and Table~\ref{tab:judge_llms_version} in the Appendix.
\subsection{Experiment 1: Comparison with Baselines}
\textbf{Setup}: We conducted experiments on four datasets, evaluating all baseline methods and our proposed MinosEval. The multilingual NLI model in our approach is mDeBERTa-v3-base-mnli-xnli.

\noindent \textbf{Result and Analysis}: The results for the two self-built datasets are presented in Table~\ref{tab:zh_main_exp} and~\ref{main_exp}, while those for the two sampled datasets are shown in Table~\ref{tab:en_main_exp}. Tables~\ref{tab:zh_factoid_exp} and \ref{tab:en_nonfactoid_exp} compare the performance of MinosEval with methods specialized for factoid and non-factoid questions. Overall, the experimental results demonstrate that our proposed approach outperforms the baseline across several evaluation metrics, showing closer alignment with human-annotated rank orders. It performs better when the questions are correctly classified. It also suggests that when the base LLM's capabilities are already strong, using a simple LLM evaluation method, such as a listwise approach, may be more cost-effective than using a specially trained smaller evaluation model. Additionally, we provide a further discussion of the performance and differences of methods to factoid questions in Section~\ref{app:factoid}.

When using the Fact detection module for automated QA classification, performance is slightly impacted but remains competitive. On the ANTIQUE\_S5 and TREC-DL-NF\_S5, our approach may underperform relative to pairwise or listwise methods when correct factoid and non-factoid classification is missing. This is due to the limited number of candidate responses (2 to 4), where a single misordering can significantly affect results. Also, the explicit qualities of the model responses in these datasets, such as length and logic, vary considerably, leading to insignificant performance differences. However, as the LLMs have evolved, this issue is less prominent in larger datasets, such as AlginBench\_Minos and GaoKaoBench\_Minos, where the number of candidate responses is greater and their quality has improved.

\begin{table}[ht]
\centering
\caption{Performance of Factoid Methods on the Factoid subset AlignBench\_Minos}
\label{tab:zh_factoid_exp}
\resizebox{0.8\columnwidth}{!}{
  \begin{tabular}{lcccc}
    \toprule
    \multirow{2}{*}{Method} & \multicolumn{4}{c}{Factoid AlignBench\_Minos.}\\
    \cmidrule(lr){2-5}
    & K & S & \makecell[c]{RBO \\ ($p$=0.5)} & \makecell[c]{RBO \\ ($p$=0.9)} \\
    \midrule
    Pointwise & 26.42  & 34.12 & 46.35 & 82.83 \\
    Pairwise & 30.08  & 38.04 & 48.52 & 83.50 \\
    Listwise & 38.95  & 48.97 & \textbf{54.50} & 85.34 \\
    FActScore & 29.54  & 36.13 & 48.48 & 83.56 \\
    RefChecker (NLIChecker) & 18.75  & 22.33 & 39.80 & 80.94 \\
    RefChecker (LLMChecker) & 32.62  & 39.76 & 48.72 & 83.69 \\
    MinosEval (LLM-based NLI) & 39.09 & 47.96 & 55.27 & 85.52 \\
    \rowcolor{blue!8}MinosEval (BERT-based NLI) & \textbf{42.77}  & \textbf{51.66}  & 54.13 & \textbf{85.67} \\
    \bottomrule
  \end{tabular}
}
\end{table}

\begin{table}[ht]
\small
\centering
\caption{Performance of Non-Factoid Methods on the Non-Factoid subset of ANTIQUE\_S5.}
\label{tab:en_nonfactoid_exp}
\resizebox{1.0\columnwidth}{!}{
  \begin{tabular}{lcccc}
    \toprule
    \multirow{2}{*}{Method} & \multicolumn{4}{c}{Non-Factoid ANTIQUE\_S5}\\
    \cmidrule(lr){2-5}
    & K & S & \makecell[c]{RBO \\ ($p$=0.5)} & \makecell[c]{RBO \\ ($p$=0.9)} \\
    \midrule
    PEDANTS & 37.84\textsubscript{$\pm$ 0.0161}  & 41.58\textsubscript{$\pm$ 0.0163} & 73.40\textsubscript{$\pm$ 0.0048} & 93.00\textsubscript{$\pm$ 0.0010} \\
    BEM & 39.13\textsubscript{$\pm$ 0.0239}  & 42.76\textsubscript{$\pm$ 0.0288} & 75.99\textsubscript{$\pm$ 0.0110} & 93.49\textsubscript{$\pm$ 0.0027} \\
    Pointwise & 49.16\textsubscript{$\pm$ 0.0131}  & 55.00\textsubscript{$\pm$ 0.0139} & 75.92\textsubscript{$\pm$ 0.0069} & 94.14\textsubscript{$\pm$ 0.0014} \\
    Pairwise & 62.83\textsubscript{$\pm$ 0.0159}  & 68.28\textsubscript{$\pm$ 0.0138} & 80.90\textsubscript{$\pm$ 0.0045} & 95.31\textsubscript{$\pm$ 0.0011} \\
    Listwise & 62.31\textsubscript{$\pm$ 0.0343}  & 68.52\textsubscript{$\pm$ 0.0311} & 82.88\textsubscript{$\pm$ 0.0154} & 95.80\textsubscript{$\pm$ 0.0039} \\
    PROMETHEUS 2 & 51.25\textsubscript{$\pm$ 0.0204}  & 57.00\textsubscript{$\pm$ 0.0211} & 76.47\textsubscript{$\pm$ 0.0077} & 94.26\textsubscript{$\pm$ 0.0015} \\
    AUTO-J & 49.54\textsubscript{$\pm$ 0.0065}  & 55.18\textsubscript{$\pm$ 0.0081} & 76.28\textsubscript{$\pm$ 0.0041} & 94.23\textsubscript{$\pm$ 0.0009} \\
    \rowcolor{blue!8}MinosEval & \textbf{71.95}\textsubscript{$\pm$ 0.0122}  & \textbf{75.85}\textsubscript{$\pm$ 0.0080}  & \textbf{86.31}\textsubscript{$\pm$ 0.0098} & \textbf{96.72}\textsubscript{$\pm$ 0.0023} \\
    \bottomrule
  \end{tabular}
}
\end{table}
\subsection{Experiment 2: Ablation Study}
\textbf{Setup}: We conducted ablation experiments on key steps, strategies, and models of MinosEval, including the classification between factoid and non-factoid questions, two ranking strategies, the necessity of key point extraction, and the LLMs and NLI models used.

\noindent \textbf{Result and Analysis}: Table~\ref{tab:classifyacc} presents the accuracy of open-ended QA classification using GPT-4o. The overall performance exceeds 90\% when appropriate demonstration examples using in-context learning. However, combining the results from Table~\ref{tab:zh_main_exp} and Table~\ref{tab:en_main_exp} reveals that this step introduces cascading errors.  Table~\ref{tab:ablationmodule} illustrates the performance of our two key strategies on AlginBench\_Minos when the distinction between factoid and non-factoid questions is not made. It is evident that the adaptive keypoint scoring approach performs poorly on non-factoid QA due to its inability to identify suitable key points. In contrast, the instance-aware Listwise ranking strategy generalizes better, leveraging key information from examples and reference answers for factoid QA. 

Table~\ref{tab:nliablation} shows the results of computing entailment probabilities directly between the reference answers and model responses, without decomposing the reference answers into key points. We also conducted a comparative experiment by swapping the reference answer and the model response as the premise and hypothesis, in order to eliminate the effect of directional bias. Furthermore, a discussion on the premise and hypothesis settings in the NLI task is provided in the appendix~\ref{app:nlidis}.
From the table, It is evident that directly comparing the NLI relationship between model outputs and reference answers is limited by the complex semantics of the sentences, which can lead to confusion. Our MinosEval addresses the issue of semantic ambiguity caused by too many key points in a single sentence, enabling more fine-grained evaluation.
Table~\ref{tab:ablationmodel} presents the performance of four LLMs and a multilingual NLI model, showing competitive results and a strong generalizability. Table~\ref{tab:overall-ranking} presents the overall model rankings produced by AlpacaEval, our MinosEval, and human annotations.

\begin{table}[ht]
\centering
\small
\caption{Accuracy of GPT-4o in Classifying Factoid and Non-Factoid Questions in Zero-shot and Few-shot Settings (n denotes the number of demonstrations).}
\resizebox{1.0\columnwidth}{!}{
    \begin{tabular}{l|cc}
    \toprule
    \textbf{Dataset} & Zero-shot & Few-shot (n=5) \\
    \midrule
    AlignBench\_Minos & 89.17 & 97.70 \\
    GaokaoBench\_Minos & 86.17 & 95.06 \\
    ANTIQUE\_S5 & 81.82 & 90.91 \\
    TREC-DL-NF\_S5 & 86.60 & 91.00 \\
    \bottomrule
    \end{tabular}
}
\label{tab:classifyacc}
\end{table}
\subsection{Experiment 3: Robustness and Cost}
\textbf{Setup}: We calculated the standard deviations of the results from five experiment sets on the sampled datasets ANTIQUE\_S5 and TREC-DL-NF\_S5. Using the AlginBench\_Minos dataset as an example, we compared the resource consumption of all baselines and our method.

\begin{figure}[ht]
    \centering
    \includegraphics[width=1.0\columnwidth]{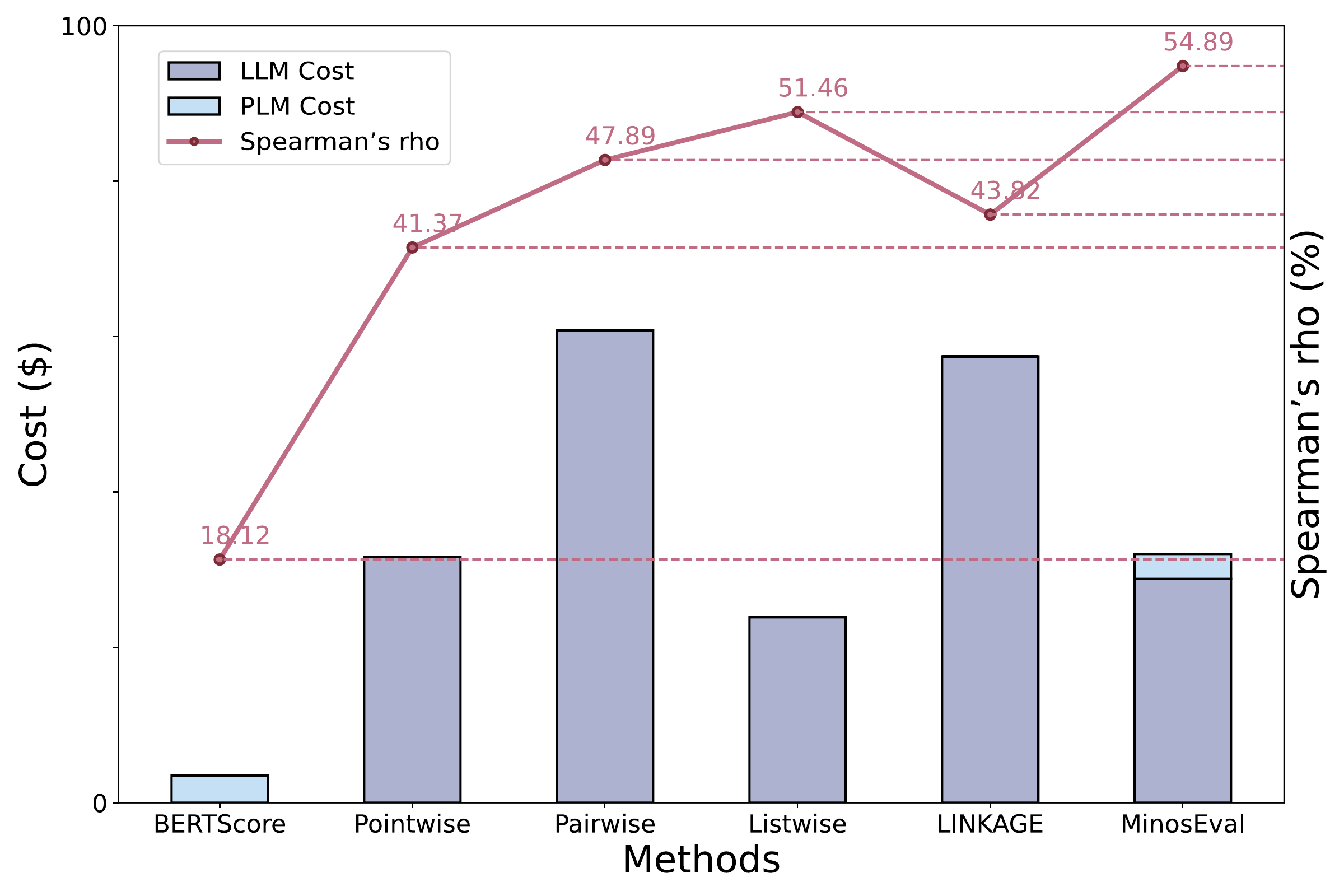}
    \caption{Comparison of Computational Cost (Scaled for Clarity) and Performance of Different Open-ended QA Evaluation Methods on AlginBench\_Minos.}
    \label{fig:cost}
\end{figure}

\noindent \textbf{Result and Analysis}: The standard deviations of all methods across the five sampled datasets are presented in Table~\ref{tab:en_main_exp}. The results show that MinosEval exhibits superior robustness. Figure~\ref{fig:cost},~\ref{fig:factcost} and~\ref{fig:nonfactcost} illustrate the cost-performance trends of each method, with our approach demonstrating a more favorable ``price/performance'' ratio.
\subsection{Case Study and Error Analysis}
Figure~\ref{fig:casestudy} provides a comprehensive, intuitive sample study to illustrate the performance of MinosEval, along with its interpretability. The upper part shows the alignment between model responses and the corresponding key points, with boxes highlighting the correct, incomplete, and incorrect points. The lower part illustrates the correspondence between the generated silver instances and the model responses.

To guide the customized use of MinosEval and the future of open-ended QA evaluation, we reviewed 200 errors using GPT-4o as basic LLM (from four datasets, with 100 factoid and 100 non-factoid samples). After manual classification, we identified five error categories: (a) QA classification errors (CE), (b) key point extraction errors (KPEE), (c) NLI entailment judgment errors (NLIE), (d) errors from low-quality instances (LIQ), and (e) errors from LLM listwise ranking (LLMRE). Examples are shown in Figure~\ref{fig:error}.
\section{Conclusion}
In conclusion, we present MinosEval, an LLM-based approach for open-ended QA evaluation. By distinguishing between factoid and non-factoid questions, MinosEval tailors evaluation strategies to suit each type, employing adaptive key-point scoring for factoid questions and instance-aware listwise ranking for non-factoid ones. We constructed two datasets with more candidate responses and greater difficulty, and conducted experiments across four datasets. The results validate the effectiveness of the individual modules, demonstrate the robustness of the method, and analyze its computational efficiency. MinosEval shows improved alignment with human annotations and offers more interpretable results. We plan to open-source this project in hopes of contributing to more effective LLM evaluation research.

\section*{Limitations}
\label{sec:con_limit}
Evaluating open-ended QA is inherently challenging, and our work faces some limitations. We did not train a specialized NLI model, instead relying on the widely-used mDeBERTa-v3-base-mnli-xnli. The performance of such generalized models may be constrained in more specialized evaluation scenarios. Additionally, the boundary between factoid and non-factoid problems can be ambiguous, with some questions potentially fitting both categories. In such cases, combining the two strategies of MinosEval may be beneficial, offering a potential direction for future research.

While we have constructed datasets like AlignBench\_Minos and GaoKaoBench\_Minos, which contain a larger number of model responses, they remain limited. We look forward to expanding these resources and leveraging more community datasets to further validate and improve our approach. Despite efforts to control for randomness (by fixing temperature and format constraints), some variability remains. We will continue to explore ways to improve our work.

\section*{Ethical Considerations}
\label{sec:ethics}
This paper presents a new open-ended questioning and answering evaluation method \textbf{MinosEval}. All of the datasets used in MinosEval are adhere to ethical guidelines and respect copyright laws. The entire data collection process is free of issues of copyright and issues of privacy, and there are three types of data sources, including license applications, the open source community, and public file cleaning and organizing. Meanwhile, the manual participation part in the dataset construction process was all done by the authors of this paper without any ethical issues.

\section*{Acknowledgments}
We would like to express our sincere gratitude to the anonymous reviewers for their valuable feedback. We also thank the Chairs and the organizing staff for their dedicated efforts in facilitating this work. This work was supported by the National Key Research and Development Program of China (Grant 2023YFF1204904).

\bibliography{anthology,custom}

\clearpage
\appendix
\section{Supplementary materials for the MinosEval experiment}
In this section, we provide supplemental experiment results in this paper. Tables~\ref{tab:ablationmodule},~\ref{tab:nliablation},~\ref{tab:ablationmodel},~\ref{tab:overall-ranking},~\ref{main_exp}, and Figures~\ref{fig:factcost},~\ref{fig:nonfactcost} complement some additional experiment results. Figure~\ref{fig:casestudy} illustrates a detailed case study.
\renewcommand{\thetable}{B\arabic{table}}
\renewcommand{\thefigure}{B\arabic{figure}}
\setcounter{figure}{0}
\setcounter{table}{0}

\begin{table}[ht]
\centering
\small
\caption{Ablation Study on key evaluation strategies of MinosEval on AlignBench\_Minos. AKPS (Adaptive Key-Point Scoring) and IALR (Instance-Aware Listwise Ranking). ``$\dagger$'' denotes manual classification of factoid and non-factoid questions.}
\resizebox{\columnwidth}{!}{
    \begin{tabular}{l|cccc}
    \toprule
    \textbf{Method} & K & S & \makecell[c]{RBO \\ ($p$=0.5)} & \makecell[c]{RBO \\ ($p$=0.9)} \\
    \midrule
    Minos\_AKPS & 24.88 & 30.03 & 45.20 & 82.49 \\
    Minos\_IALR & 44.32 & 54.01 & \textbf{58.43} & 86.54 \\
    MinosEval & 45.28 & 54.89 & 56.30 & 86.28 \\
    MinosEval$^\dagger$ & \textbf{47.68} & \textbf{57.38} & 57.09 & \textbf{86.62} \\
    \bottomrule
    \end{tabular}
}
\label{tab:ablationmodule}
\end{table}

\begin{table}[ht]
\centering
\small
\caption{Direct Comparison of NLI Relations Between Model Responses (R) and Reference Answers (A) on Factoid AlignBench\_Minos.}
\resizebox{\columnwidth}{!}{
    \begin{tabular}{l|cccc}
    \toprule
    \textbf{Method} & K & S & \makecell[c]{RBO \\ ($p$=0.5)} & \makecell[c]{RBO \\ ($p$=0.9)} \\
    \midrule
    NLI (R,A) & 17.59 & 21.76 & 40.99 & 81.15 \\
    NLI (A,R) & 18.17 & 22.73 & 40.47 & 81.06 \\
    MinosEval & \textbf{42.77} & \textbf{51.66} & \textbf{54.13} & \textbf{85.67} \\
    \bottomrule
    \end{tabular}
}
\label{tab:nliablation}
\end{table}

\begin{table}[ht]
\centering
\small
\caption{Ablation Study on LLM and NLI model of MinosEval used on AlignBench\_Minos.}
\resizebox{\columnwidth}{!}{
    \begin{tabular}{l|cccc}
    \toprule
    \textbf{Setting} & K & S & \makecell[c]{RBO \\ ($p$=0.5)} & \makecell[c]{RBO \\ ($p$=0.9)} \\
    \midrule
    GPT-3.5+mDeBERTa & 40.65 & 49.94 & 54.30 & 85.55 \\
    GPT-3.5+mDeBERTa$^\dagger$ & 37.57 & 47.17 & 53.07& 85.11 \\
    \midrule
    Qwen2.5+mDeBERTa & 40.32 & 49.56 & 53.25 & 85.33 \\
    Qwen2.5+mDeBERTa$^\dagger$ & 42.57 & 51.67& 55.02 & 85.85 \\
    \midrule
    DeepSeekV2.5+mDeBERTa & 43.39 & 52.74 & 56.23 & 86.21 \\
    DeepSeekV2.5+mDeBERTa$^\dagger$ & 44.50 & 53.73 & 56.39 & 52.28 \\
    \midrule
    GPT-4o+XLM-RoBERTa & 43.67 & 52.58 & 56.08 & 86.11 \\
    GPT-4o+XLM-RoBERTa$^\dagger$ & 46.33 & 55.52 & \textbf{57.72} & 86.56 \\
    \midrule
    GPT-4o+mDeBERTa & 45.28 & 54.89 & 56.30 & 86.28 \\
    GPT-4o+mDeBERTa$^\dagger$ & \textbf{47.68} & \textbf{57.38} & 57.09 & \textbf{86.62} \\
    \bottomrule
    \end{tabular}
}
\label{tab:ablationmodel}
\end{table}

\begin{table*}[ht]
\centering
\small
\caption{Comparison of Overall Model Rankings from AlpacaEval, MinosEval, and Human Annotations on the AlignBench\_Minos Dataset.}
\resizebox{\textwidth}{!}{
    \begin{tabular}{l|llllll}
    \toprule
    \textbf{Method} & rank1 & rank2 & rank3 & rank4 & rank5 & rank6 \\\midrule
   AlpacaEval & GPT-4o & Qwen2.5 & ChatGLM4 & InternLM2 & GPT-3.5 & LLAMA3.1 \\
    MinosEval & GPT-4o & ChatGLM4 & Qwen2.5 & GPT-3.5 & InternLM2 & LLAMA3.1\\
   Manually annotated & GPT-4o & ChatGLM4 & Qwen2.5 & InternLM2 & GPT-3.5 & LLAMA3.1\\
    \bottomrule
    \end{tabular}
}
\label{tab:overall-ranking}
\end{table*}

\begin{figure}[ht]
    \centering
    \includegraphics[width=1.0\columnwidth]{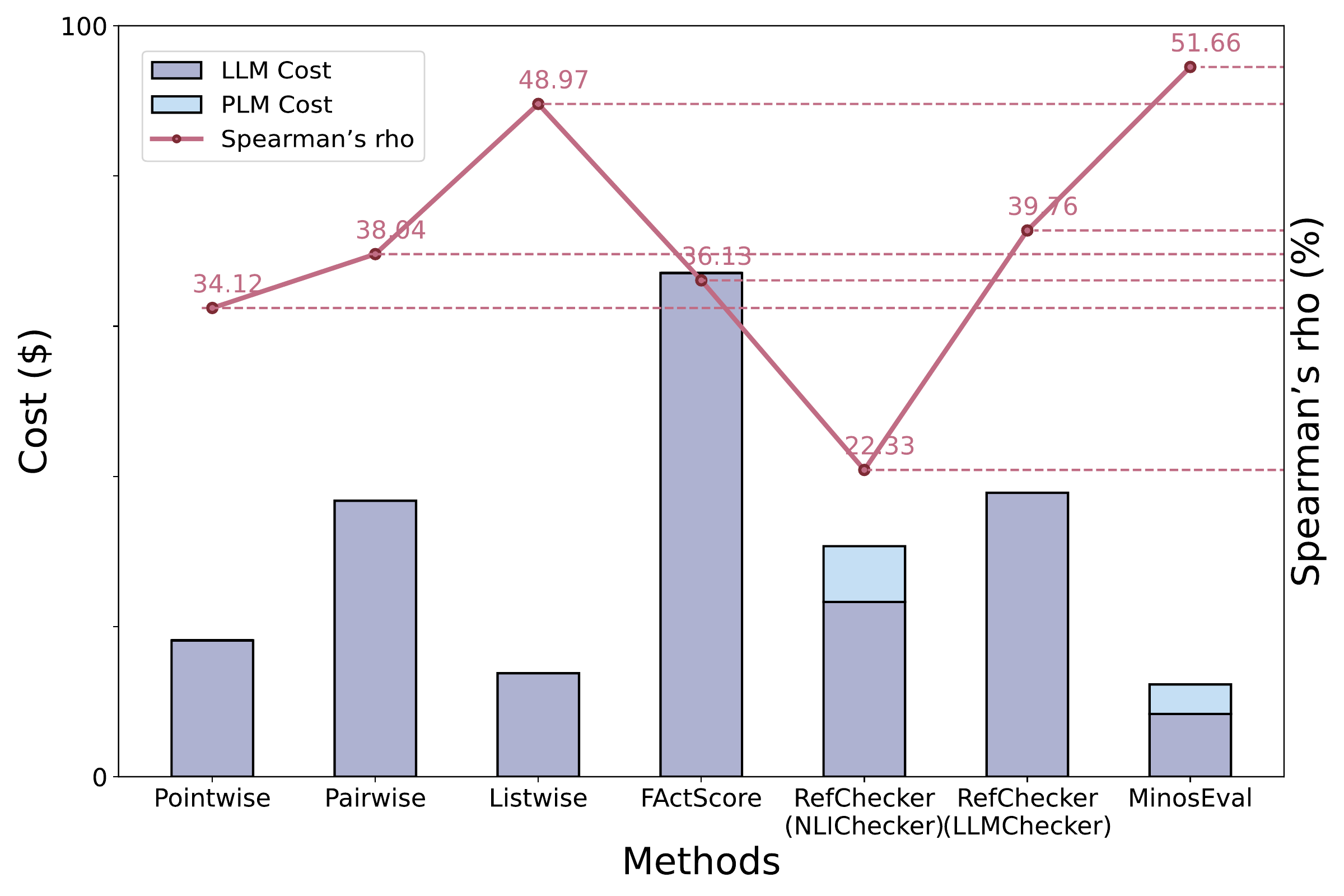}
    \caption{Comparison of Computational Cost (Scaled for Clarity) and Performance of Different Open-ended QA Evaluation Methods on Factoid AlginBench\_Minos.}
    \label{fig:factcost}
\end{figure}

\begin{figure}[ht]
    \centering
    \includegraphics[width=1.0\columnwidth]{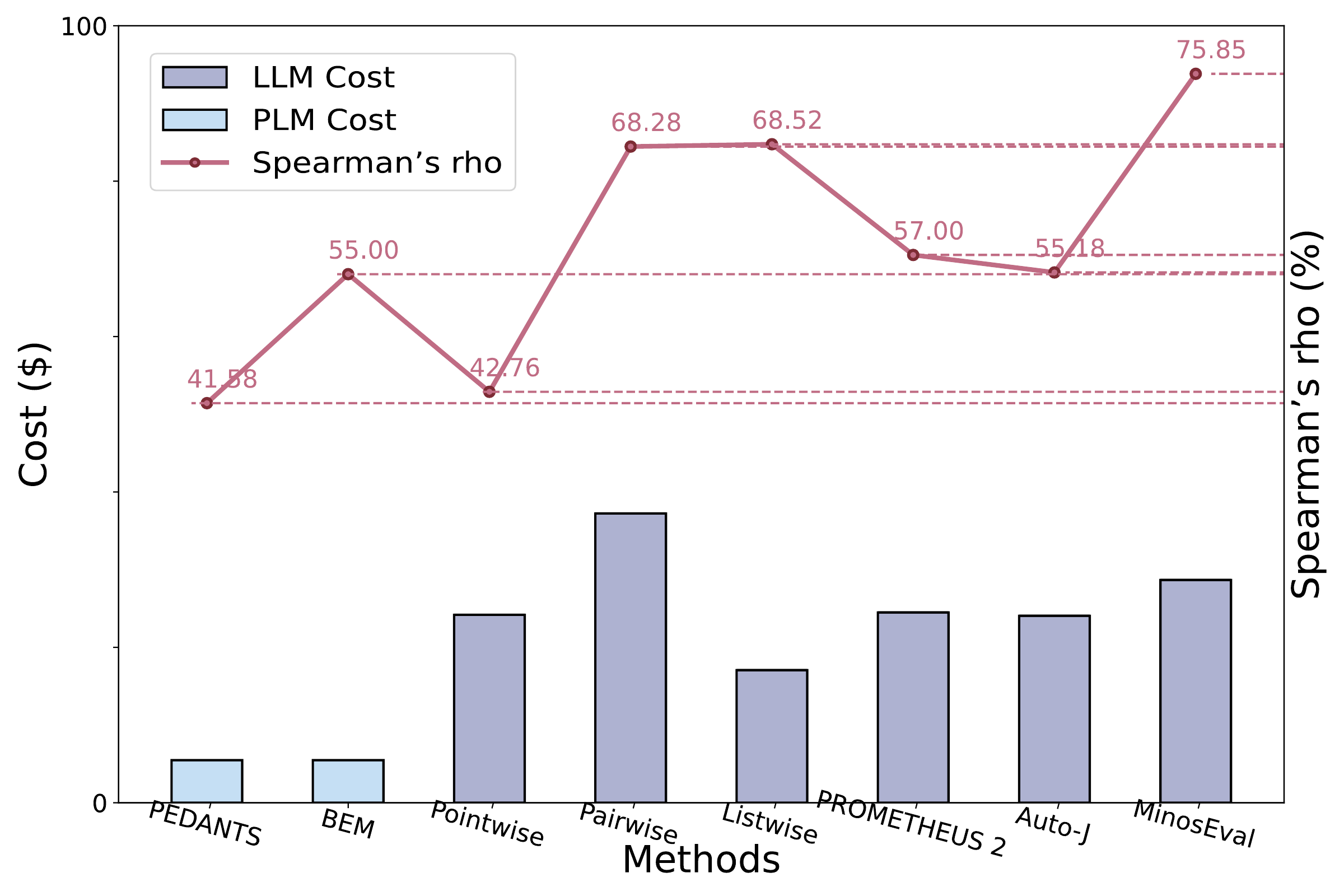}
    \caption{Comparison of Computational Cost (Scaled for Clarity) and Performance of Different Open-ended QA Evaluation Methods on Non-Factoid ANTIQUE\_S5.}
    \label{fig:nonfactcost}
\end{figure}

\begin{figure}[ht]
    \centering
    \includegraphics[width=1.0\columnwidth]{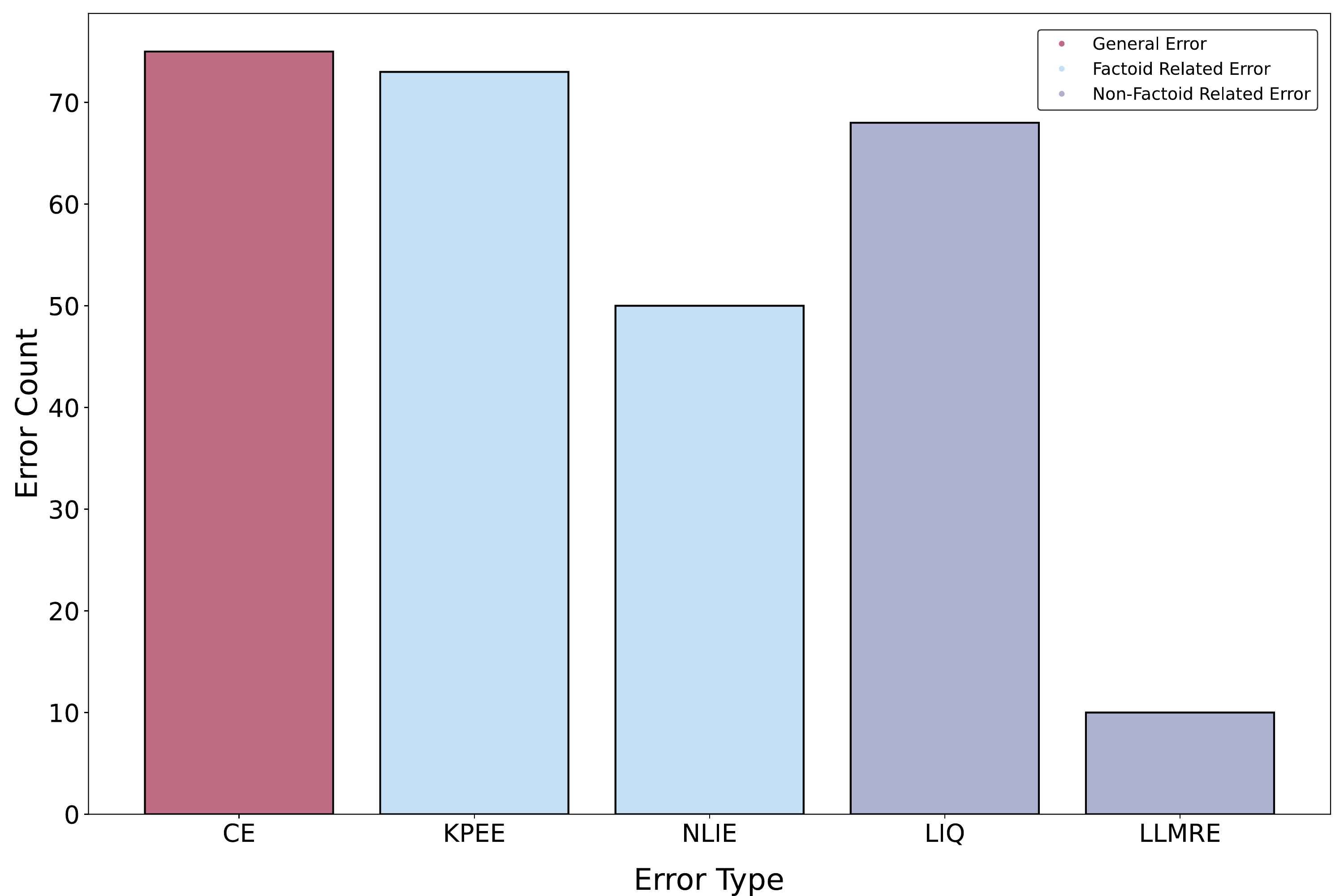}
    \caption{Error Analysis of MinosEval on 200 Samples: CE represents QA classification errors, KPEE represents key point extraction errors, NLIE represents NLI model entailment judgment errors, LIQ represents errors caused by low-quality instances, and LLMRE represents errors from LLM listwise ranking.}
    \label{fig:error}
\end{figure}

\newpage

\begin{figure*}[t]
    \centering
    \includegraphics[width=1.0\linewidth]{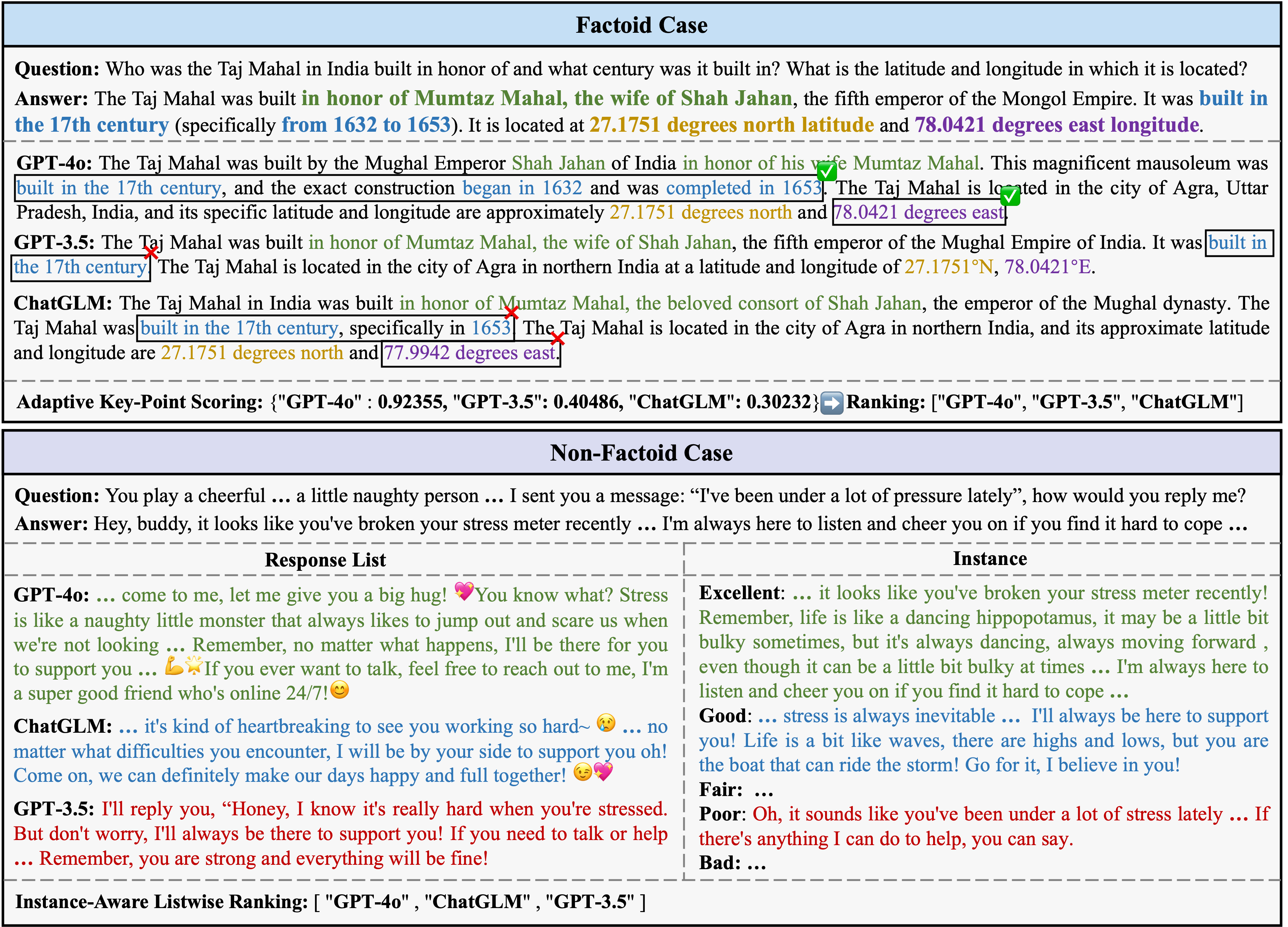}
    \caption{A case study on factoid and non-factoid open-ended QA of MinosEval.}
    \label{fig:casestudy}
\end{figure*}

\begin{table*}[ht]
\caption{Performance of Methods on AlignBench\_Minos (Fctoid / Non-Factoid) and GaokaoBnech\_Minos (Fctoid / Non-Factoid): Kendall's Tau (K), Spearman’s Rho (S), and Rank-biased Overlap (RBO) for $p$=0.5 and $p$=0.9. ``$\dagger$'' denotes manual classification of factoid and non-factoid questions.}
\label{main_exp}
\resizebox{\textwidth}{!}{
  \begin{tabular}{clcccccccc}
    \toprule
    & \multirow{2}{*}{Method} & \multicolumn{4}{c}{AlignBench\_Minos (Factoid / Non-factoid)} & \multicolumn{4}{c}{{GaokaoBench\_Minos} (Factoid / Non-factoid)} \\
    \cmidrule(lr){3-6} \cmidrule(lr){7-10}
    & & K & S & \makecell[c]{RBO \\ ($p$=0.5)} & \makecell[c]{RBO \\ ($p$=0.9)} & K & S & \makecell[c]{RBO \\ ($p$=0.5)} & \makecell[c]{RBO \\ ($p$=0.9)} \\
    \midrule
    \multirow{2}{*}{\makecell[c]{\text{Automatic Metrics}}}
    & ROUGE-L & 3.14 / 21.35 & 3.83 / 26.70 & 34.69 / 40.45 & 78.71 / 81.16 & 6.08 / 4.13 & 6.44 / 5.40 & 34.95 / 35.27 & 78.89 / 78.99 \\
    & BLEU & 8.58 / 20.73 & 10.86 / 26.18 & 33.59 / 38.30 & 78.88 / 80.84 & 13.50 / 9.53  & 16.24 / 10.68 & 36.48 / 36.97 & 79.97 / 79.71 \\\midrule
    \multirow{1}{*}{\makecell[c]{\text{LM-based Metrics}}}
    & BERTScore & 13.04 / 14.06 & 16.81 / 19.14 & 37.67 / 37.92 & 79.95 / 80.31 & 12.91 / 12.93 & 15.95 / 16.76 & 38.01 / 41.63 & 80.12 / 81.01 \\
    \midrule
    \multirow{3}{*}{\makecell[c]{\text{LLM} \\ \text{Evaluation}}} 
    & Pointwise & 26.42 / 38.02 & 34.12 / 47.01 & 46.35 / 48.23 & 82.83 / 84.18 & 24.64 / 35.17 & 30.63 / 43.62 & 38.84 / 45.07 & 81.09 / 83.28 \\
    & Pairwise & 30.08 / 45.35 & 38.04 / 55.57 & 48.52 / 53.97 & 83.50 / 85.90 & 33.59 / 53.47 & 40.51 / 64.16 & 46.97 / 63.24 & 83.53 / 88.31 \\
    & Listwise & 38.95 / 43.44 & 48.97 / 53.39 & 54.50 / 54.56 & 85.34 / 85.86 & 38.40 / 66.15 & 45.90 / 76.66 & 51.71 / 67.90 & 85.01 / 90.49 \\
    \midrule
    \multirow{1}{*}{\makecell[c]{\text{LINKAGE}}} 
    & LINKAGE & 30.66 / 39.72 & 38.12 / 48.26 & 51.26 / 54.31  & 84.16 / 85.45 & 25.40 / 45.16 & 31.46 / 52.01 & 43.68 / 58.09 & 82.27 / 86.51 \\
    \midrule
    \multirow{2}{*}{\makecell[c]{\text{Ours}}} 
    & MinosEval & 42.77 / 47.38  & 51.66 / 58.18 & 54.13 / 58.43 & 85.67 / 86.54 & 36.40 / \textbf{71.57} & 46.20 / 81.12 & 50.47 / 73.46 & 84.51 / \textbf{91.93} \\
    & \cellcolor{blue!8}MinosEval$^\dagger$ & \cellcolor{blue!8}\textbf{44.66 / 49.90}  & \cellcolor{blue!8}\textbf{53.77 / 60.09} & 
    \cellcolor{blue!8}\textbf{54.83 / 58.67} & 
    \cellcolor{blue!8}\textbf{85.94 / 87.11}  & \cellcolor{blue!8}\textbf{41.60} / 71.39 & \cellcolor{blue!8}\textbf{53.24 / 81.21} & \cellcolor{blue!8}\textbf{52.33 / 73.54} & \cellcolor{blue!8}\textbf{85.29} / 91.89 \\
    \bottomrule
  \end{tabular}
}
\end{table*}

\clearpage
\section{Supplementary materials for the implementation details of MinosEval}
\renewcommand{\thetable}{C\arabic{table}}
\renewcommand{\thefigure}{C\arabic{figure}}
\setcounter{figure}{0}
\setcounter{table}{0}
\subsection{Overall description}
\label{app:rank_process}
\begin{figure*}[ht]
    \centering
    \includegraphics[width=1.0\textwidth]{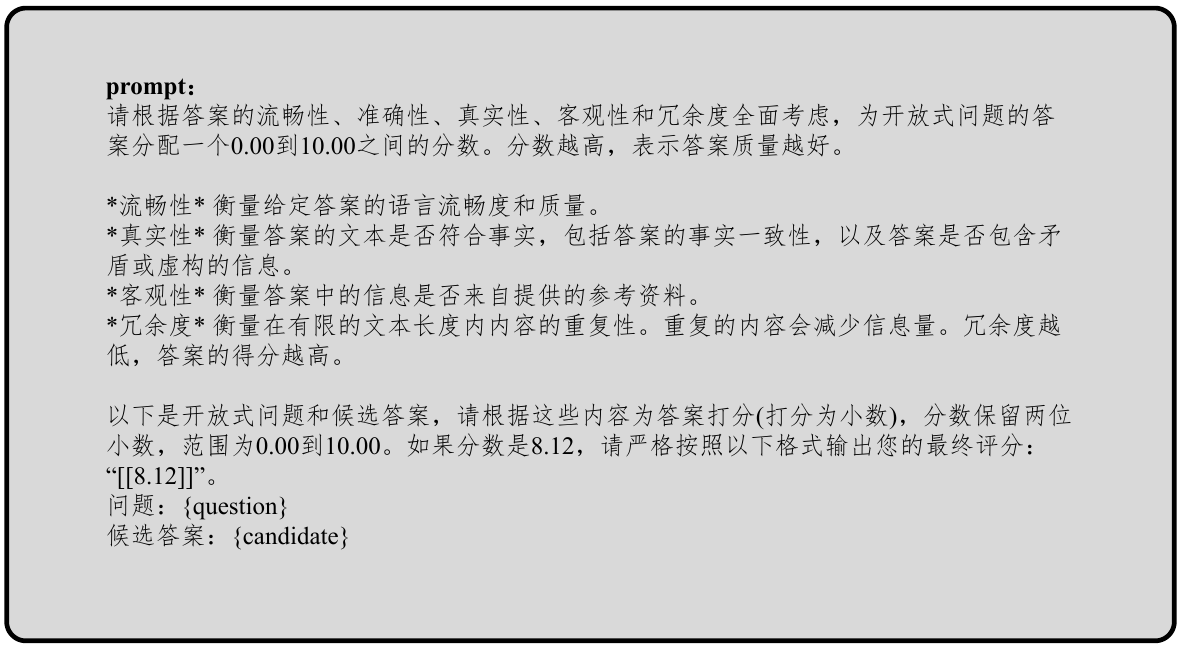}
    \caption{Prompt of pointwise method.}
    \label{fig:point_wise}
\end{figure*}

\begin{figure*}[ht]
    \centering
    \includegraphics[width=1.0\textwidth]{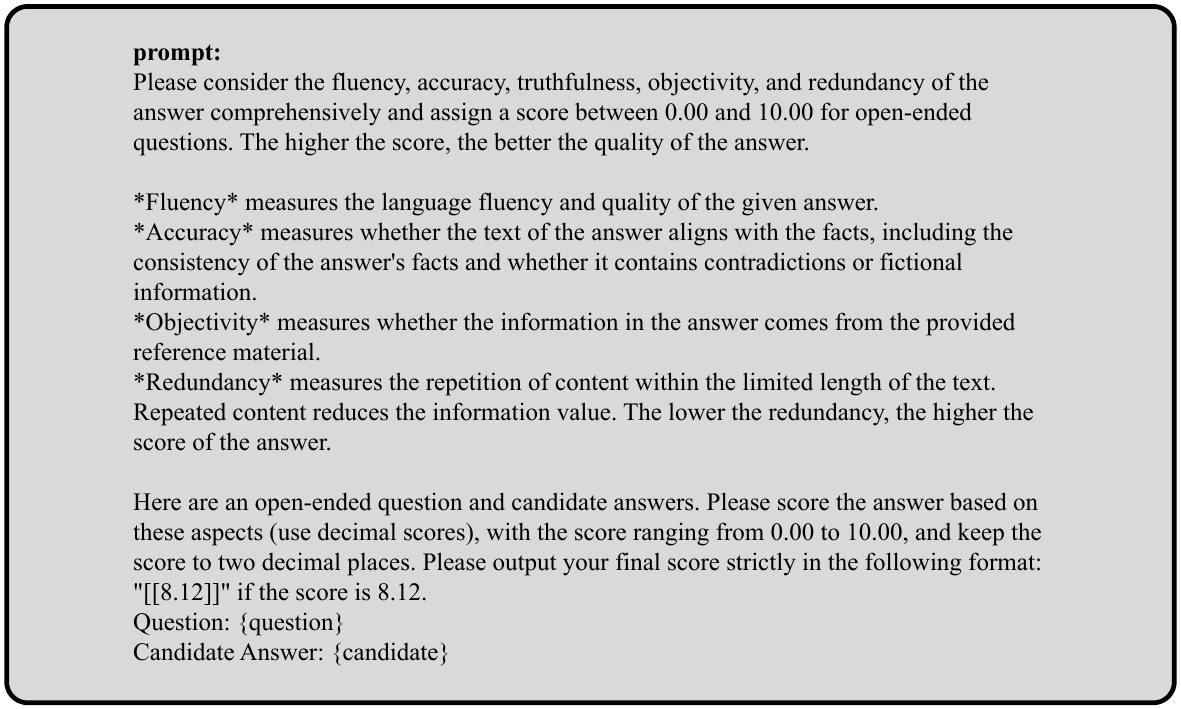}
    \caption{English version prompt of pointwise method.}
    \label{fig:point_wise_en}
\end{figure*}

\begin{figure*}[ht]
    \centering
    \includegraphics[width=1.0\textwidth]{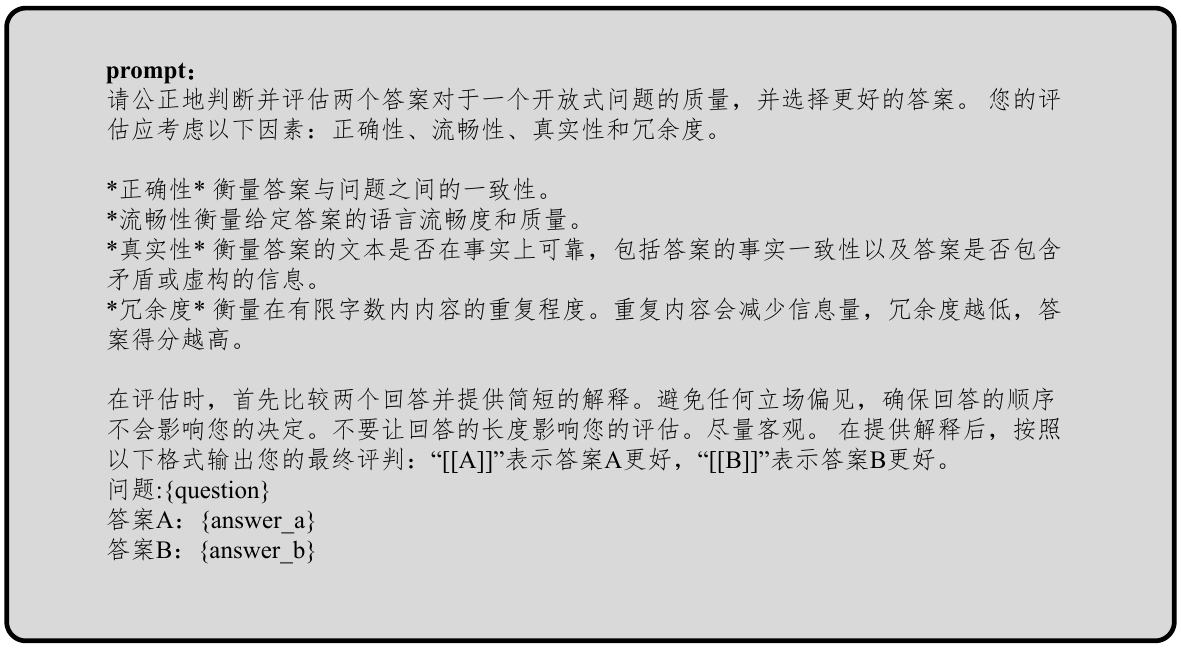}
    \caption{Prompt of pairwise method.}
    \label{fig:pair_wise}
\end{figure*}

\begin{figure*}[ht]
    \centering
    \includegraphics[width=1.0\textwidth]{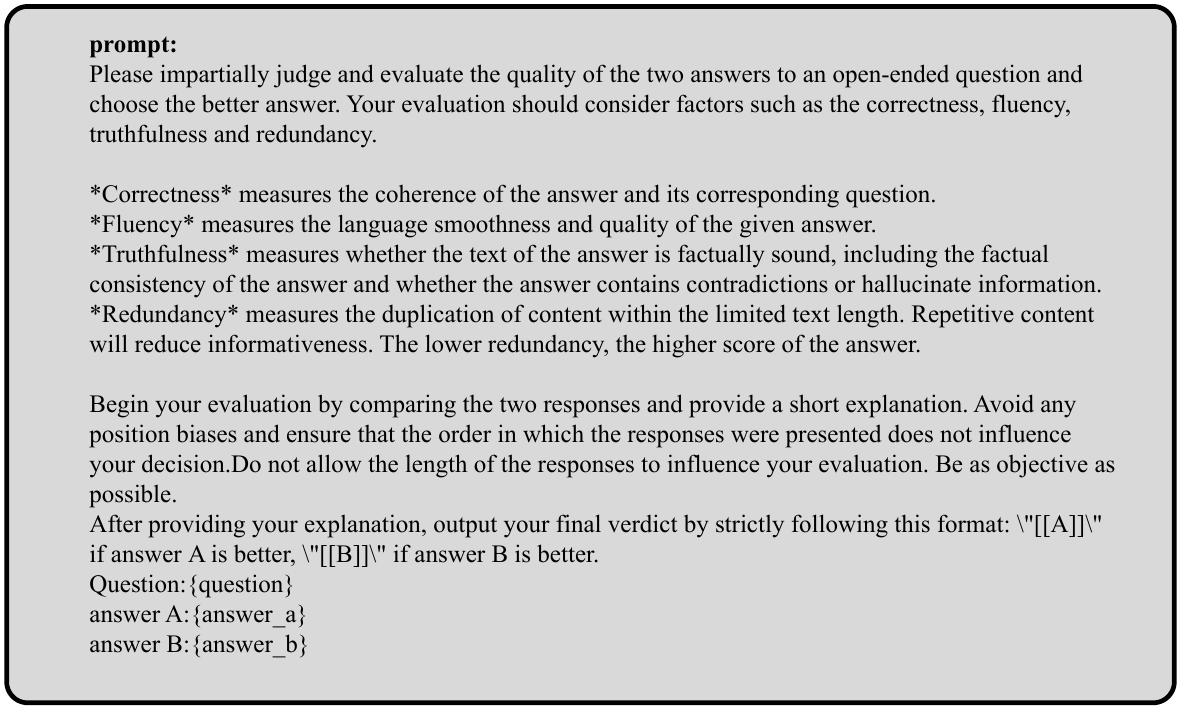}
    \caption{English version prompt of pairwise method.}
    \label{fig:pair_wise_en}
\end{figure*}

\begin{figure*}[ht]
    \centering
    \includegraphics[width=1.0\textwidth]{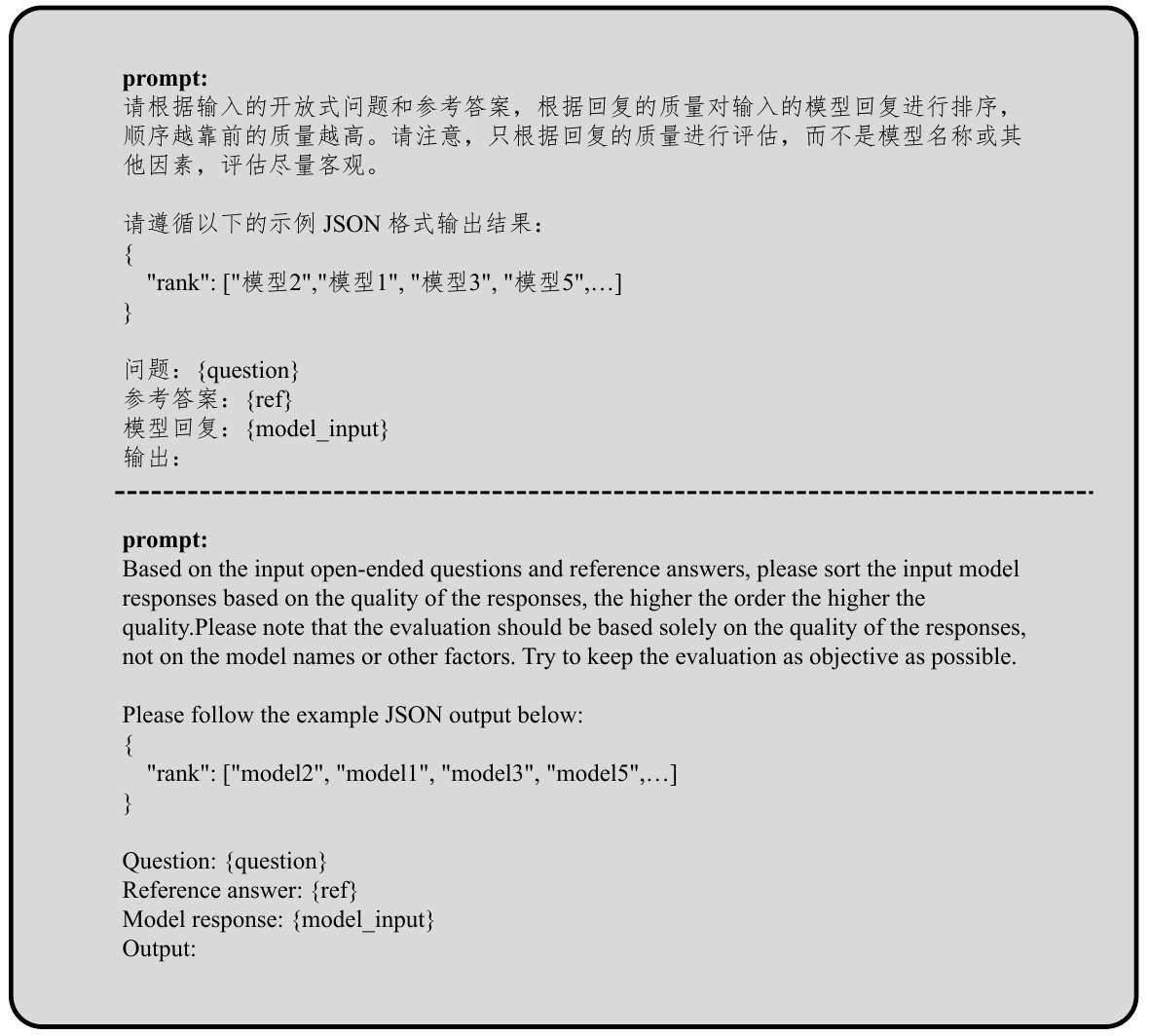}
    \caption{Prompt of listwise method.}
    \label{fig:list_wise}
\end{figure*}

\begin{figure*}[ht]
    \centering
    \includegraphics[width=1.0\textwidth]{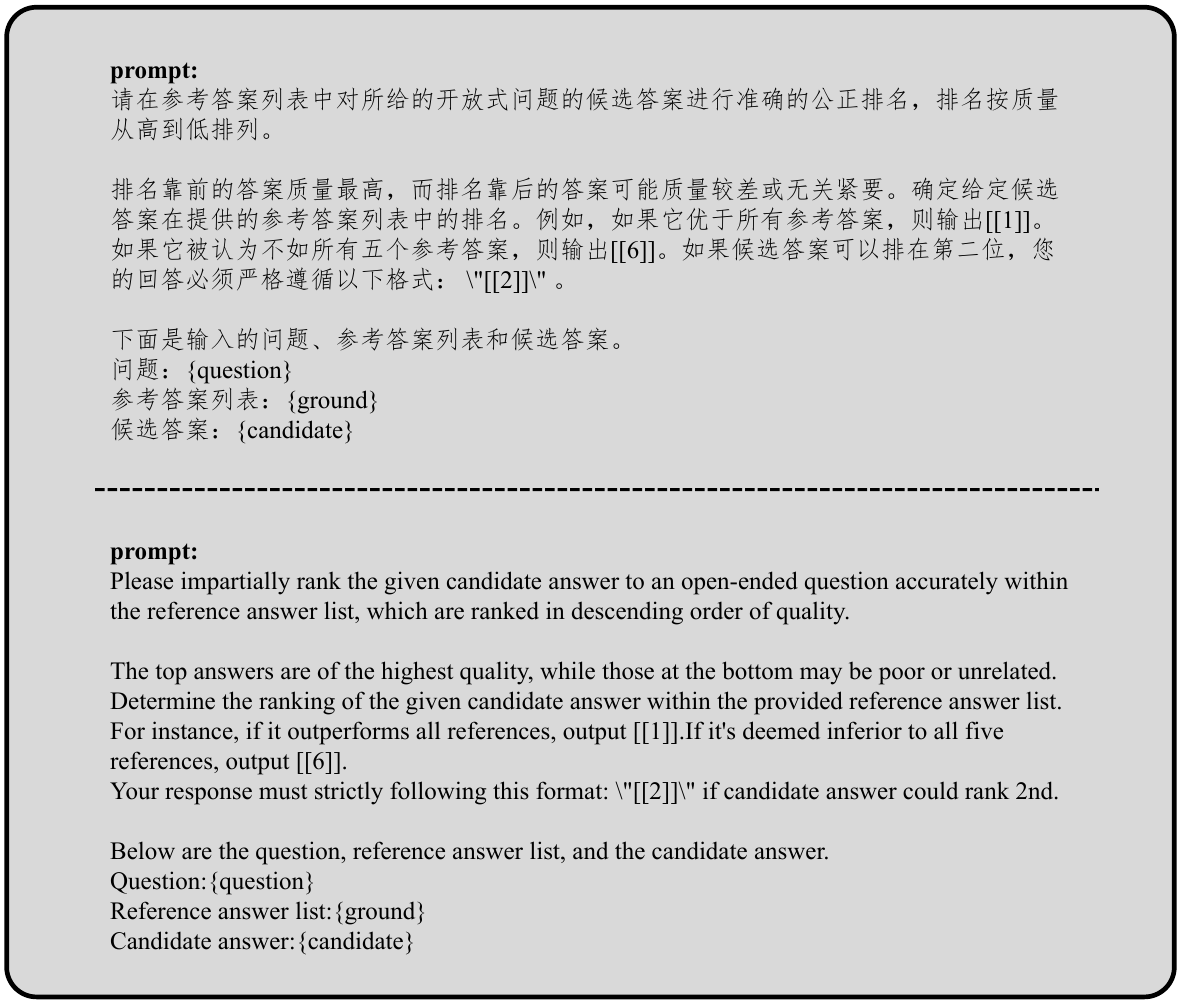}
    \caption{Prompt of linkage method.}
    \label{fig:linkage}
\end{figure*}

\begin{figure*}[ht]
    \centering
    \includegraphics[width=1.0\textwidth]{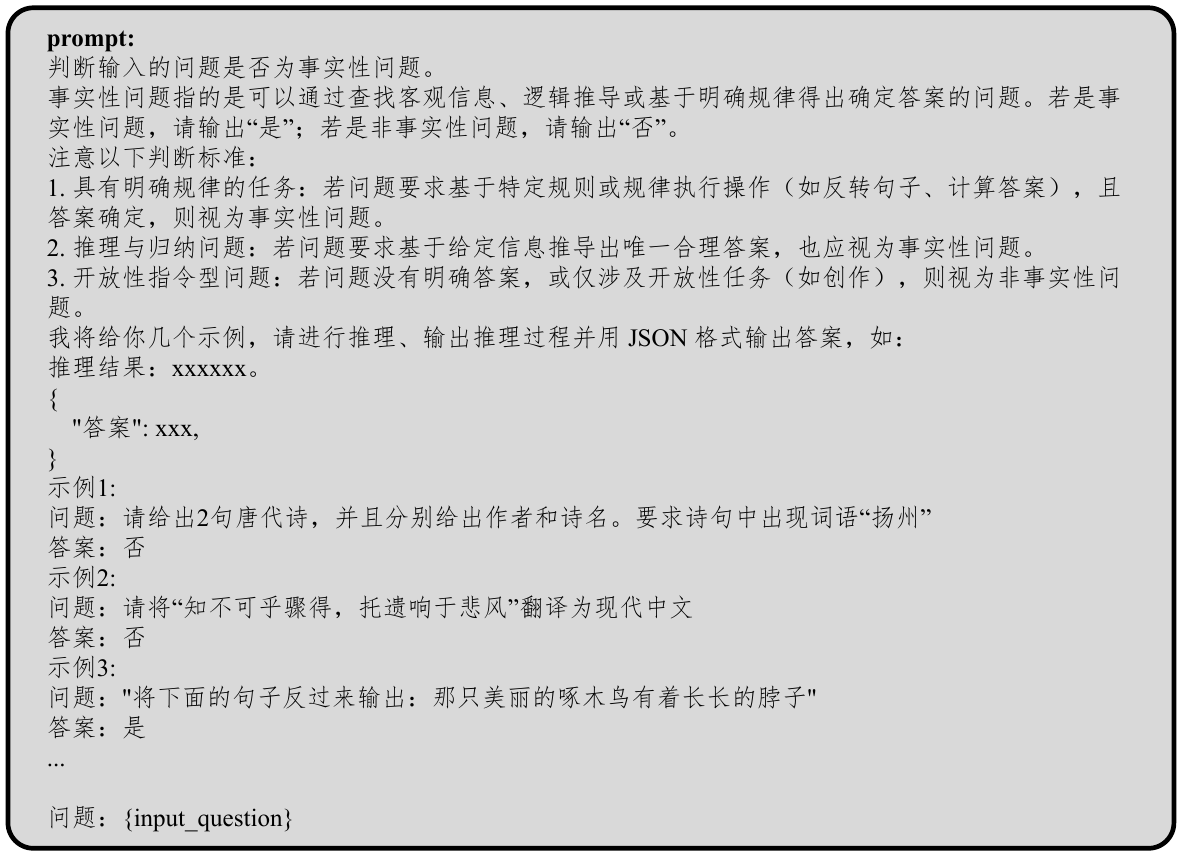}
    \caption{The prompt of factoid and non-factoid judgment in LLMs.}
    \label{fig:llmdisting_zh}
\end{figure*}

\begin{figure*}[ht]
    \centering
    \includegraphics[width=1.0\textwidth]{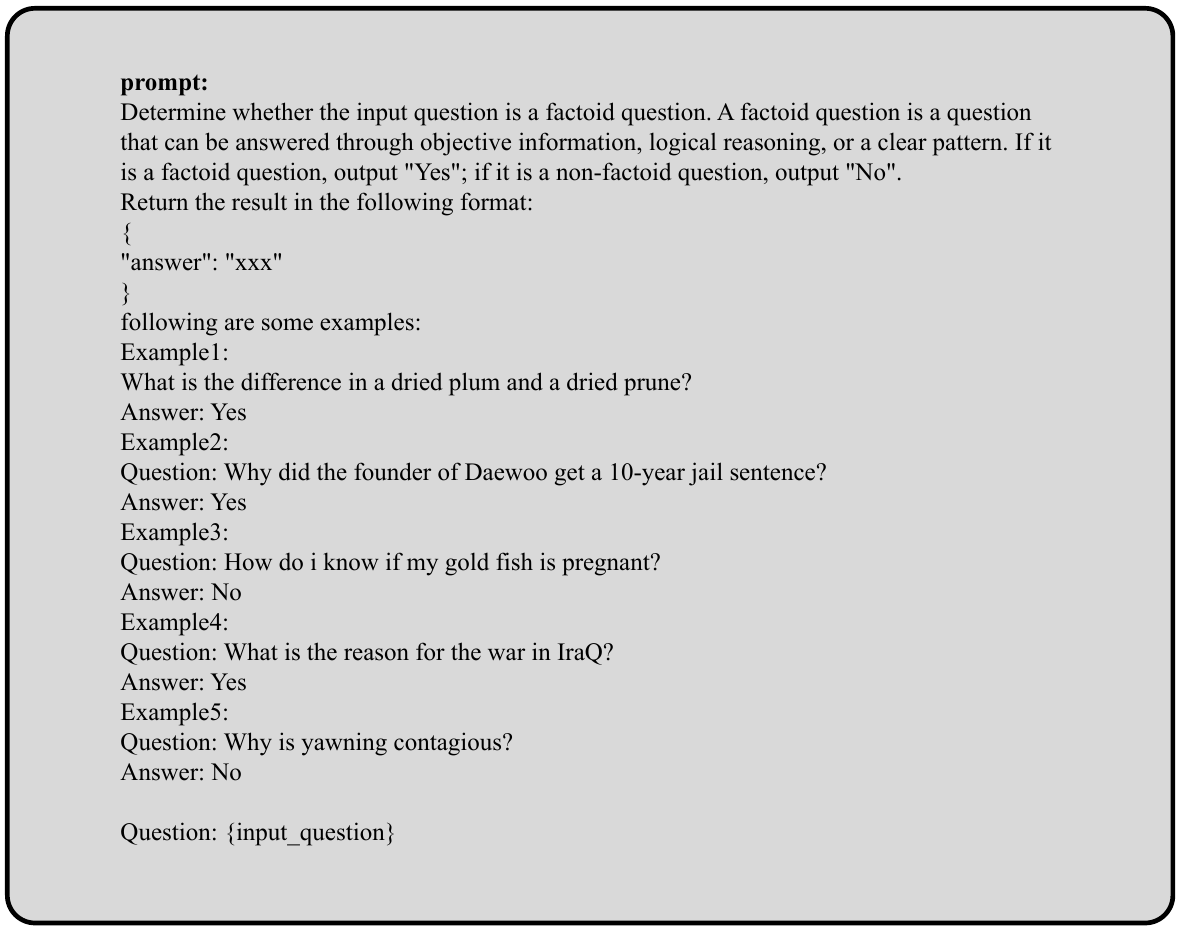}
    \caption{The english version prompt of factoid and non-factoid judgment in LLMs.}
    \label{fig:llmdisting_en}
\end{figure*}

\begin{figure*}[ht]
    \centering
    \includegraphics[width=1.0\textwidth]{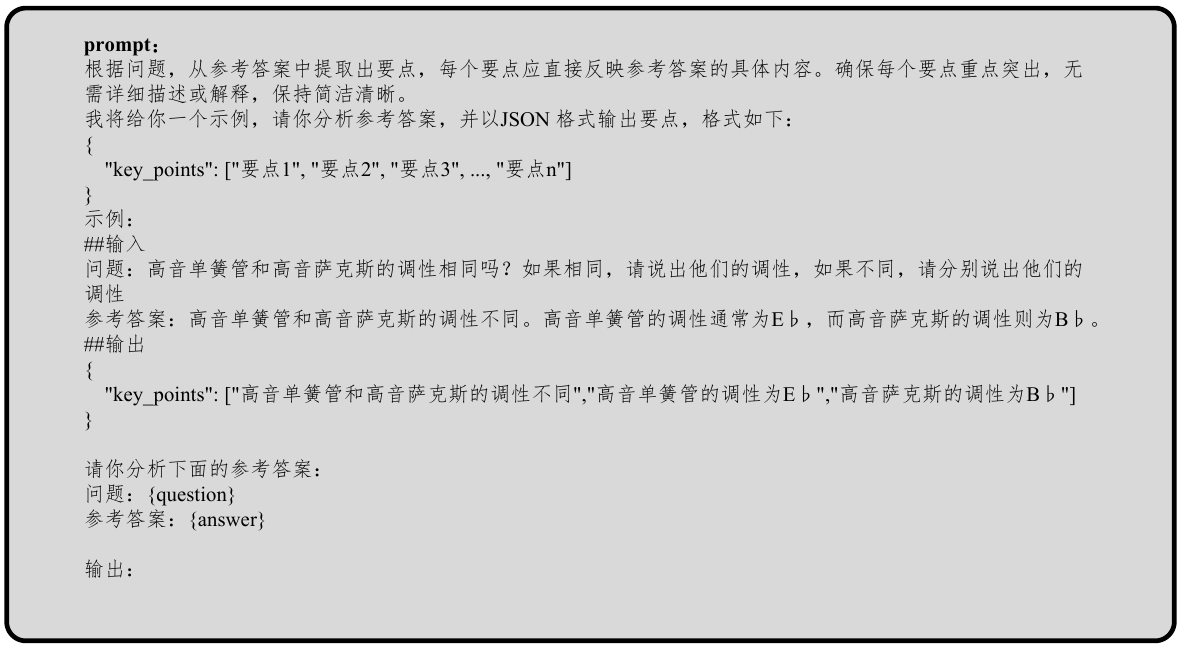}
    \caption{The prompt of LLMs to generate key points.}
    \label{fig:keypoints_zh}
\end{figure*}

\begin{figure*}[ht]
    \centering
    \includegraphics[width=1.0\textwidth]{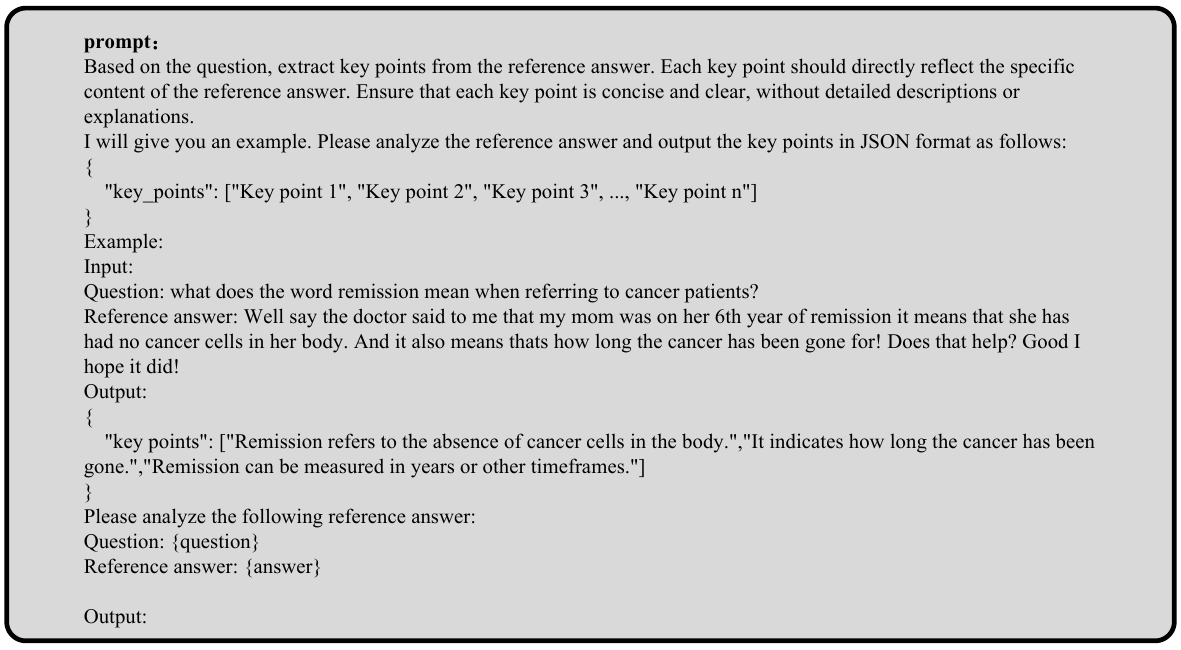}
    \caption{The english version prompt of LLMs to generate key points.}
    \label{fig:keypoints_en}
\end{figure*}

\begin{figure*}[ht]
    \centering
    \includegraphics[width=1.0\textwidth]{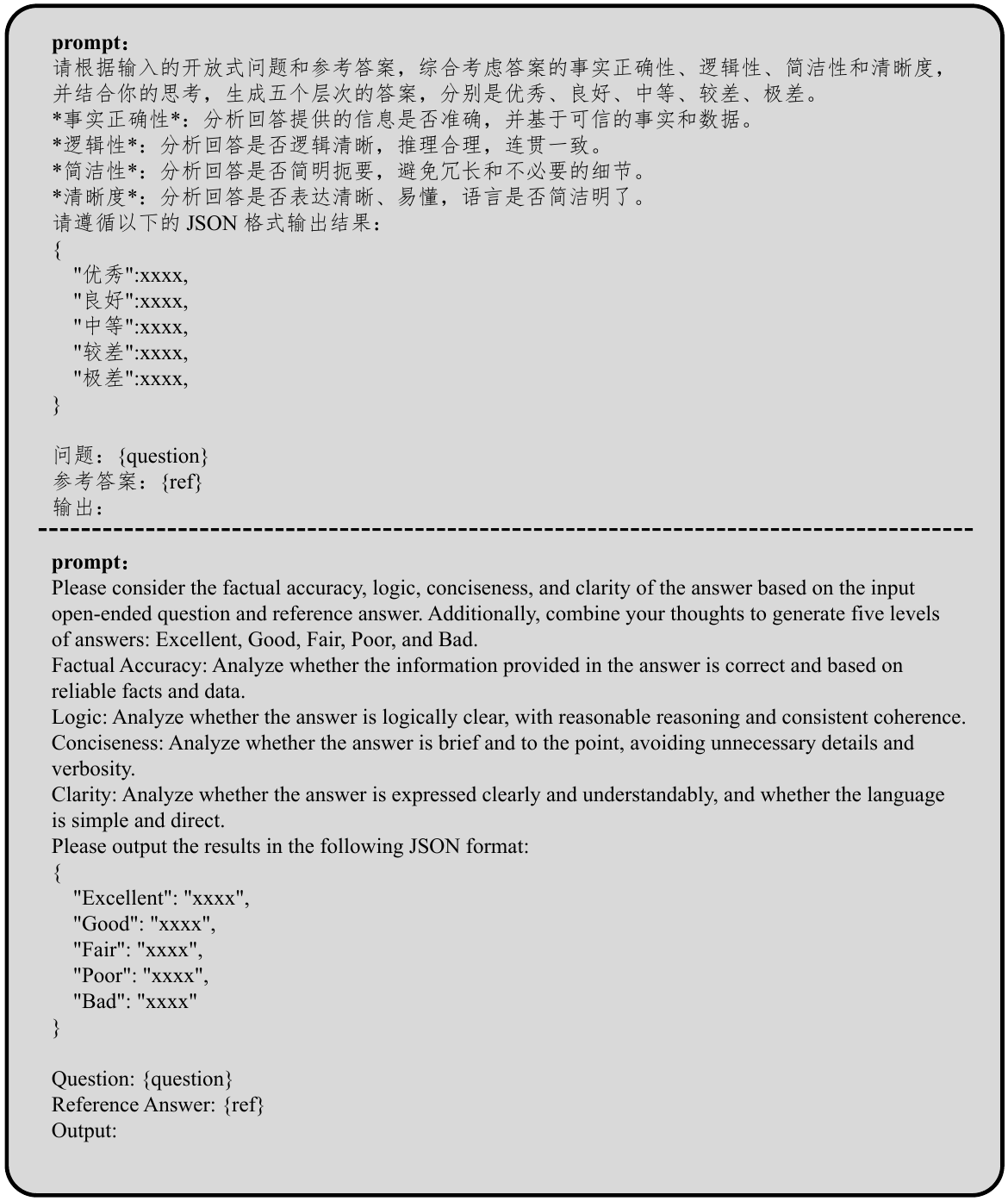}
    \caption{The prompt of LLMs to generate instance.}
    \label{fig:instance}
\end{figure*}

\begin{figure*}[ht]
    \centering
    \includegraphics[width=1.0\textwidth]{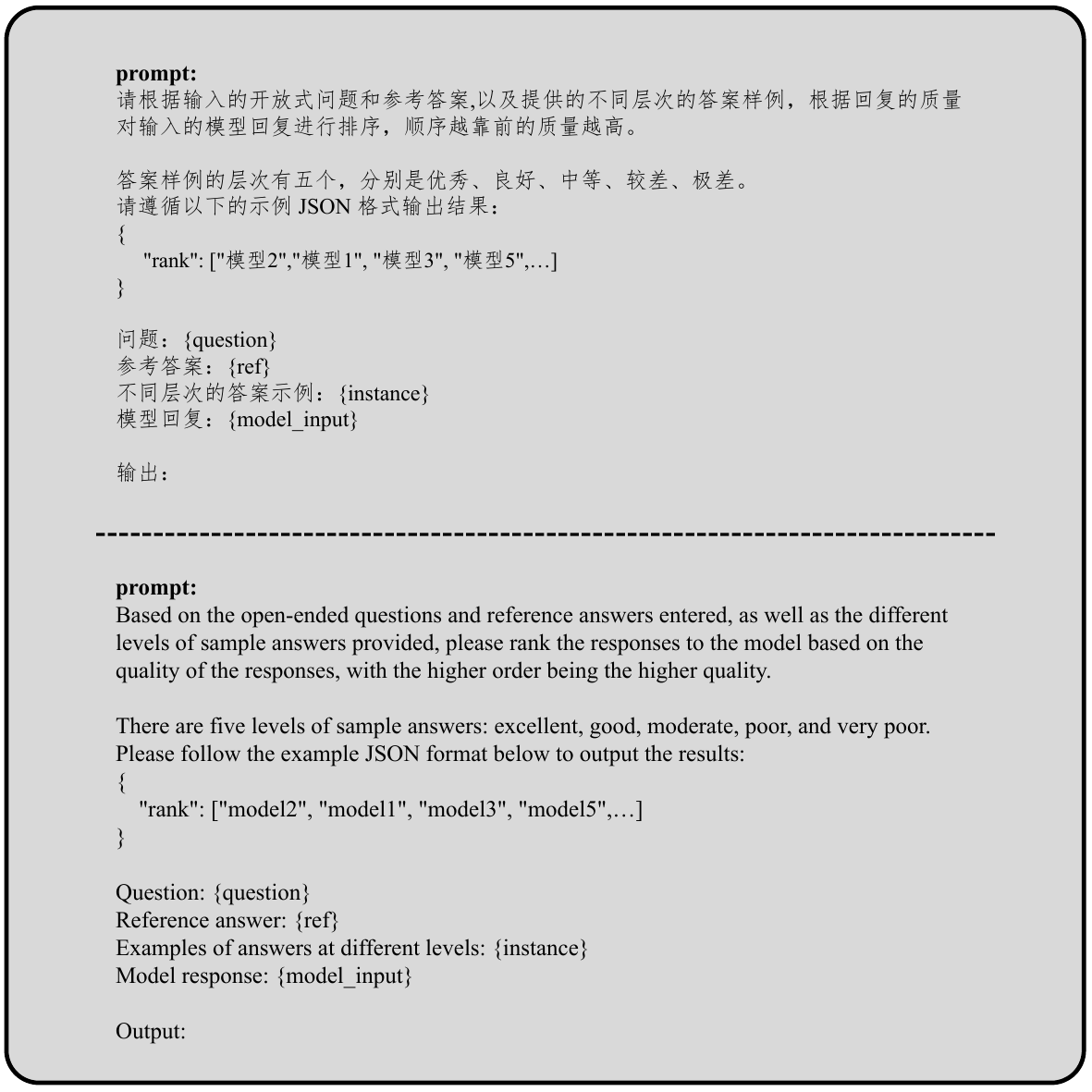}
    \caption{The prompt of LLMs to generate ranking based on instances.}
    \label{fig:list_wise_instance}
\end{figure*}

\begin{figure*}[ht]
    \centering
    \includegraphics[width=1.0\textwidth]{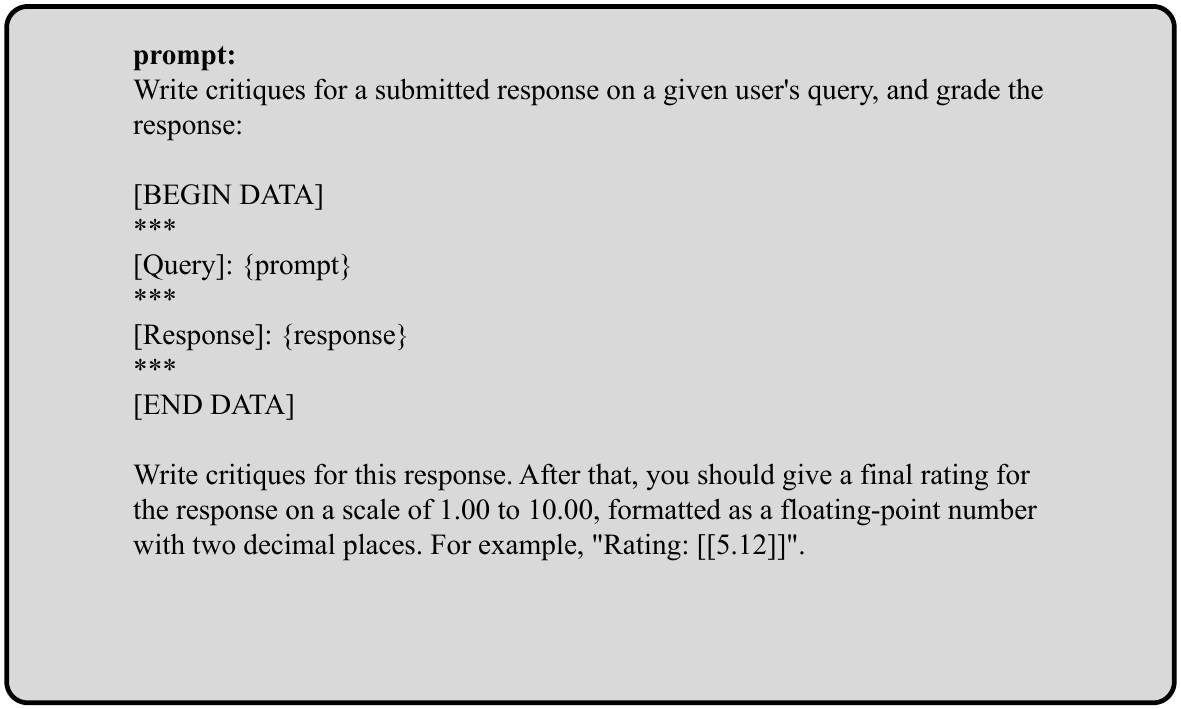}
    \caption{The prompt of AUTO-J method.}
    \label{fig:auto-j}
\end{figure*}

\begin{figure*}[ht]
    \centering
    \includegraphics[width=1.0\textwidth]{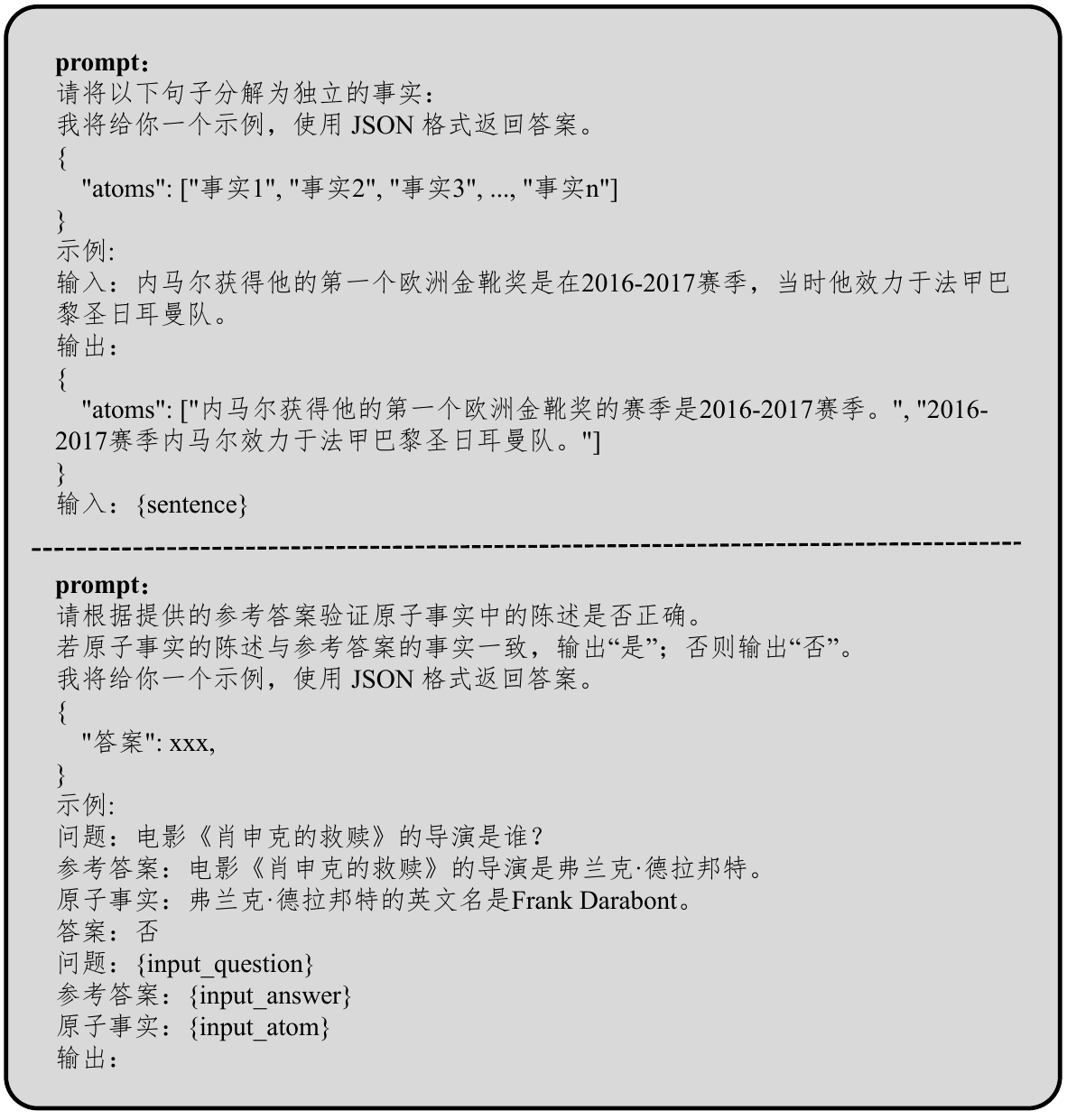}
    \caption{The prompt of FACTSCORE method.}
    \label{fig:factscore}
\end{figure*}

\begin{figure*}[ht]
    \centering
    \includegraphics[width=1.0\textwidth]{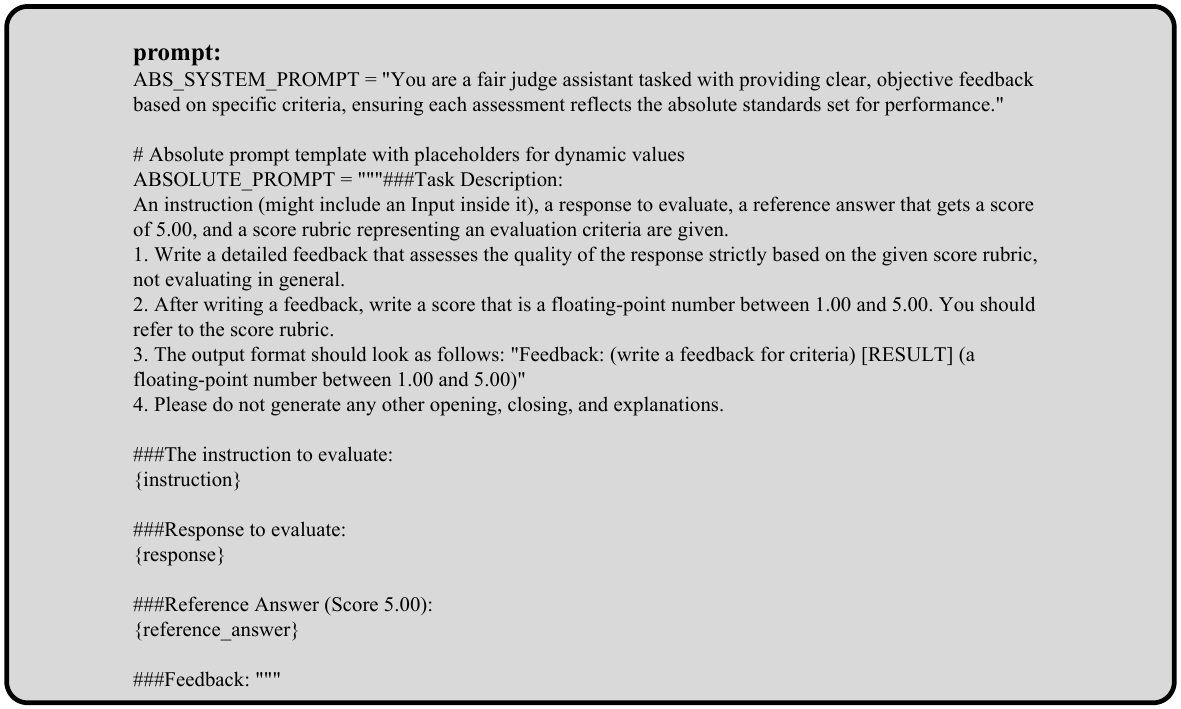}
    \caption{The prompt of PROMETHEUS 2 method.}
    \label{fig:pro}
\end{figure*}

\begin{figure*}[ht]
    \centering
    \includegraphics[width=1.0\textwidth]{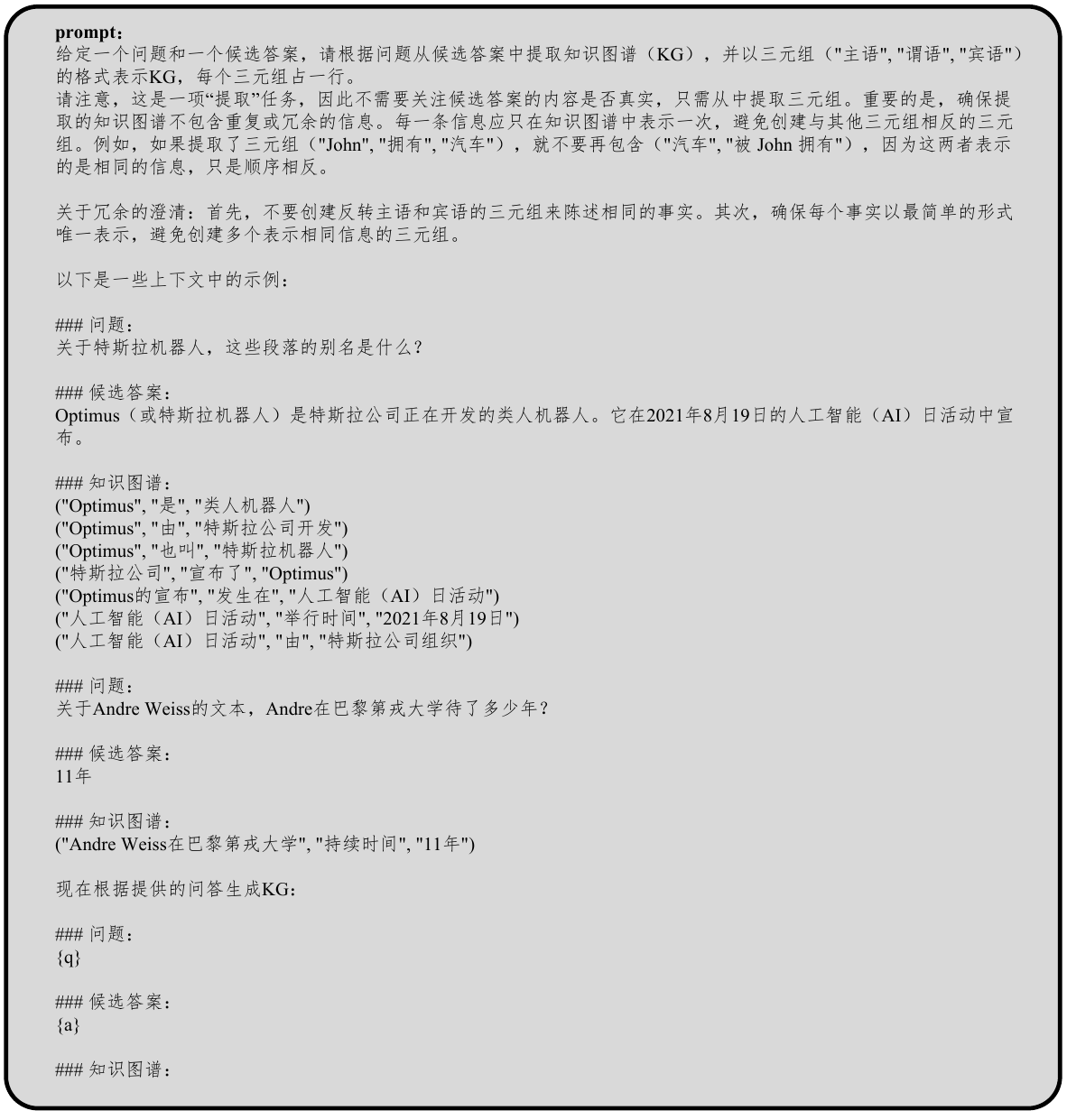}
    \caption{The prompt of refchecker method.}
    \label{fig:refchecker-extracker}
\end{figure*}

\begin{figure*}[ht]
    \centering
    \includegraphics[width=1.0\textwidth]{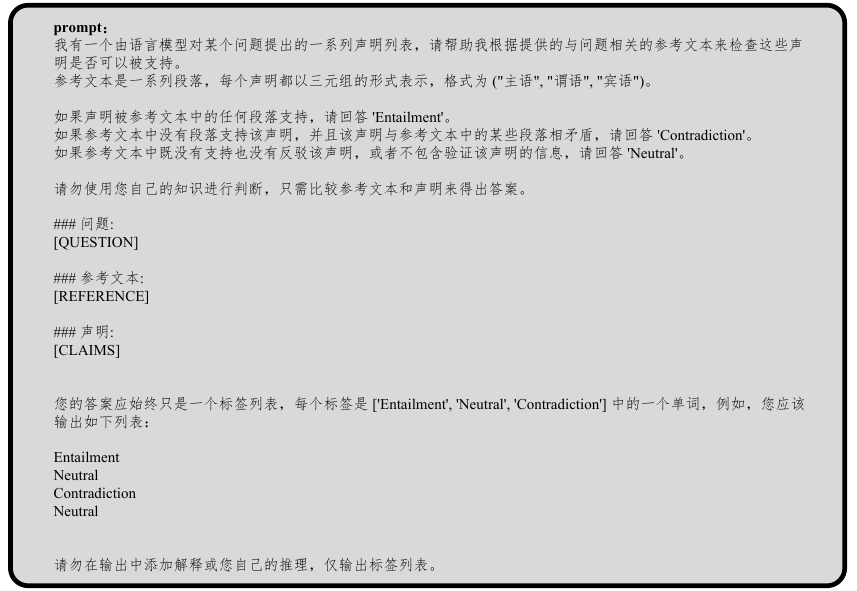}
    \caption{The prompt of refchecker method.}
    \label{fig:refchecker-checker}
\end{figure*}

In this section, we provide supplemental implementation details in this paper. We provide the specific versions of LLMs we used in this paper in Table~\ref{tab:llms_version} and Table~\ref{tab:judge_llms_version}. We list the inputs and outputs of all the methods, as well as the process of obtaining the rankings, in Table~\ref{tab:input_output}. We present the average response lengths for each ranking level under both manually annotated and MinosEval predicted rankings based on the dataset AlignBench\_Minos, as shown in Table~\ref{tab:length-bias}. Table~\ref{tab:trec_antique} lists examples of these being reclassified according to the annotation rules for factoid and non-factoid questions. Additionally, we show all the specific prompts of all evaluation methods from Figure~\ref{fig:point_wise} to Figure~\ref{fig:refchecker-checker}.

\subsection{Handling tie-breaking situations}
\label{app:tie-breaking}
It is worth noting that to ensure fair comparisons, we have minimized randomness and implemented measures to handle tie-breaking situations. For methods that use LLMs for direct scoring, including Pointwise, PROMETHEUS 2, and Auto-J, we ensure that the models output scores with two decimal precision to enhance differentiation. In cases of ties generated by the Pairwise method, we independently re-compared the responses of the tied models to determine their relative rankings. 

FactScore was proposed to evaluate the ability of LLMs to generate biographies of individuals. In this paper, we are aimed at evaluating the ability of LLMs on the open-ended QA task. Therefore, when reproducing FActScore, we directly use labeled reference answers as knowledge, rather than retrieving the relevant content from Wikipedia. In addition, since FactScore calculates whether the facts split out of each response are true or not, the result is a bunch of ``True or False'' labels. In order to eliminate the tie-breaking effect, we considered the percentage of facts, the number of facts, and the difference between the number of facts and the number of not-facts in turn to get the ranking of individual model responses.

When reproducing RefChecker, we translated the built-in prompts into Chinese to fit the Chinese dataset. We evaluated two types of Checkers, and similarly to MinosEval, the LLMChecker uses GPT-4o as the base LLM, while the NLIChecker uses mDeBERTa as the base model. In addition, since RefChecker gets the entailment relationship between the fact triplets extracted from each response and the reference answer, it gets a bunch of ``Entailment, Neutral, and Contradiction'' labels, which are then aggregated to get the corresponding scores. In order to eliminate the tie-breaking effect, we consider the difference between the entailment score and the contradiction score, the entailment score, the difference between the number of entailments and the number of contradictions, and the number of entailments in order to obtain the ranking of the individual model responses.

When reproducing the PROMETHEUS 2 and Auto-J methods on non-factoid questions, we provided two evaluation approaches: Direct Scoring and Pairwise Ranking. Given that Pairwise Ranking is highly resource-intensive, we chose to use the Direct Scoring approach.

\begin{table}[ht]
\centering
\small
\caption{Specific LLMs version for generating responses.}
\resizebox{\columnwidth}{!}{
    \begin{tabular}{l|l}
    \toprule
    \textbf{Model name} & Version \\\midrule
    GPT-4o & gpt-4o-2024-08-06 \\
    GPT-3.5 & gpt-3.5-turbo-0613 \\
    Qwen & qwen2.5-7B-Instruct \\
    ChatGLM & glm-4-9b-chat \\
    LLAMA & llama-3.1-8B-Instruct \\
    InternLM & internlm2-chat-7b \\
    \bottomrule
    \end{tabular}
}
\label{tab:llms_version}
\end{table}

\begin{table}[ht]
\centering
\small
\caption{Specific LLMs version in the ablation study.}
\resizebox{\columnwidth}{!}{
    \begin{tabular}{l|l}
    \toprule
    \textbf{Model name} & Version \\\midrule
    GPT-4o & gpt-4o-2024-08-06 \\
    GPT-3.5 & gpt-3.5-turbo-0613 \\
    Qwen & qwen2.5-72b-instruct \\
    DeepSeek & DeepSeek-V2.5-1210 \\
    \bottomrule
    \end{tabular}
}
\label{tab:judge_llms_version}
\end{table}
\subsection{Discussion on Factoid evaluation methods}
\label{app:factoid}
For these factoid QA evaluation methods FactScore, RefChecker, and the Adaptive Key-Point Scoring in MinosEval (MinosEval\_AKPS) that we propose, we provide a more refined discussion. Although all are motivated by the same underlying goal, they differ in their implementation approach.

\textbf{FactScore} evaluates the quality of a model response by extracting descriptive fragments from the response, getting relevant knowledge from a specific knowledge base (e.g., Wikipedia), and then determining whether these fragments are facts based on the retrieved or provided knowledge.

\textbf{RefChecker} evaluates the quality of a model response by extracting fact triplets and assessing their entailment with the reference answer using a Checker, which can be either an LLM or an NLI model. The final score is obtained by aggregating the entailment between these triples and the reference answer to reflect the quality of the responses.

In contrast, \textbf{MinosEval\_AKPS} takes a different approach by analyzing the key information in the reference answer, extracting multiple key points, and then determining whether each model response contains these key points to calculate a score.

From a performance point of view, it is more efficient to extract key facts from reference answers than to analyze individual model responses during the evaluation process. The latter faces the problem of variable quality of responses, as lower quality responses may introduce noise terms that affect the entailment judgment.

Additionally, the strategy and prompt for extracting key facts need to be carefully designed. For example, FactScore uses the strategy of breaking down sentences and then decomposing facts, which may lead to semantically repetitive items. For the final scoring, both FactScore and RefChecker get one-hot labels, i.e., ``True or False'', `` Entailment, Neutral, and Contradiction''. Therefore, a more customized score aggregation strategy may be needed to avoid a tie.

Furthermore, extracting key points only once from the reference answer helps reduce computational costs, especially when the number of model responses is large. This strategy offers better cost-effectiveness.

\subsection{Discussion on bias related to answer length}
\label{app:length-bias}
Length bias presents a common challenge in classic LLM-as-a-judge approaches. To examine whether this bias exists in our setting, based on the dataset AlignBench\_Minos, We conducted a comparative analysis of the average response length between manual annotation and MinosEval predicted at different ranking positions. The average length of the reference answers is also provided. From the results, the responses from the models are all longer than the reference answers, which is a regular phenomenon at LLM at the moment.
Responses to open-ended QA depend to a certain extent on length, Longer responses will contain more valuable information, shorter responses may be missing information. The effect of length may not be significant at moderate quality rankings, such as Rank3 and Rank4 for factoid questions, Rank2 and Rank3 for non-factoid questions.

\subsection{Discussion on the premise and hypothesis settings in the NLI task}
\label{app:nlidis}
Natural Language Inference (NLI) is the task of identifying the logical relationship between a premise (A) and a hypothesis (B). It categorizes this relationship into three labels: entailment (B can be logically inferred from A), contradiction (B directly contradicts A), and neutral (B is neither supported nor contradicted by A). Formally, NLI aims to determine the semantic and logical relation between sentence pairs. 

In our strategy for factoid questions, the directionality between premises and hypotheses is critical to ensure valid logical inference. By decomposing reference answers into key points for fine-grained assessment, we observe that model responses (as premises) can reliably entail key points (as hypotheses), whereas the reverse direction fails due to incomplete contextual information in key points. For instance, the response "Li Hua is walking on the street, and his dog Huang is running around" entails the key points "Li Hua is walking on the street" and "Huang is running around", but the key points alone cannot reconstruct the full response without additional context (e.g., "Huang" referring to "dog Huang"). This asymmetry highlights the importance of using model outputs as premises rather than key points to avoid spurious neutrality or contradictions, ensuring that NLI labels reflect meaningful logical relationships grounded in complete information.

\begin{table*}[ht]
\centering
\caption{The input, post-processing, and output of both the baseline methods and our MinosEval.}
\renewcommand{\arraystretch}{1.2}
\small
\begin{tabular}{l|p{0.24\textwidth}|p{0.46\textwidth}}
\toprule
\textbf{Method} & \textbf{Input} & \textbf{Post-processing} \\
\midrule
BLEU & \makecell[l]{reference answer,\\ candidate model responses} & \makecell[l]{Calculating the BLEU score\\ and ranking based on it.} \\
ROUGE-1 & \makecell[l]{reference answer,\\ candidate model responses} & \makecell[l]{Calculating the ROUGE-1 score\\ and ranking based on it.} \\
ROUGE-2 & \makecell[l]{reference answer,\\ candidate model responses} & \makecell[l]{Calculating the ROUGE-2 score\\ and ranking based on it.} \\
ROUGE-L & \makecell[l]{reference answer,\\ candidate model responses} & \makecell[l]{Calculating the ROUGE-L score\\ and ranking based on it.} \\
BERTScore & \makecell[l]{reference answer,\\ candidate model responses} & \makecell[l]{Calculating the BERTScore\\ and ranking based on it.} \\
\midrule
Pointwise & \makecell[l]{question, reference answer,\\ candidate model responses,\\ dimension description} & \makecell[l]{Scoring by an LLM based on multiple dimensions\\ and ranking based on it.} \\
\midrule
Pairwise & \makecell[l]{question, reference answer,\\ candidate model response pair,\\ dimension description} & \makecell[l]{Calculating the win rate\\ and ranking based on it.} \\
\midrule
Listwise & \makecell[l]{question, answer,\\ candidate model responses} & \makecell[l]{Ranking by an LLM.} \\
\midrule
LINKAGE & \makecell[l]{question, answer,\\ candidate model responses,\\ examples of different levels} & \makecell[l]{Rating by an LLM\\ and ranking based on it.} \\
\midrule
AUTO-J & \makecell[l]{question, answer,\\ candidate model responses} & \makecell[l]{Scoring by a fine-tuned LLM\\ and ranking based on it.} \\
\midrule
PROMETHEUS 2 & \makecell[l]{question, answer,\\ candidate model responses} & \makecell[l]{Scoring by a fine-tuned LLM\\ and ranking based on it.} \\
\midrule
FActScore & \makecell[l]{question, answer,\\ candidate model responses} & \makecell[l]{(a) Breaking the response into atomic facts using an LLM.\\
(b) Determining the correctness of each atomic fact.\\
(c) Calculating accuracy and ranking based on it.} \\
\midrule
RefChecker (LLMChecker) & \makecell[l]{question, answer,\\ candidate model responses} & \makecell[l]{(a) Decomposing response into claim-triplets using an LLM.\\
(b) Verifying triplets against the answer with an LLM.\\
(c) Aggregating results and ranking.} \\
\midrule
RefChecker (NLIChecker) & \makecell[l]{question, answer,\\ candidate model responses} & \makecell[l]{(a) Decomposing response into claim-triplets using an LLM.\\
(b) Verifying triplets with NLI model.\\
(c) Aggregating results and ranking.} \\
\midrule
MinosEval\_AKPS & \makecell[l]{question, answer,\\ candidate model responses} & \makecell[l]{(a) Extracting key points from answers.\\
(b) Calculating NLI scores with responses and ranking.} \\
\midrule
MinosEval\_IALR & \makecell[l]{question, answer,\\ candidate model responses} & \makecell[l]{(a) Generating sliver instances of different levels.\\
(b) LLM-based listwise ranking.} \\
\bottomrule
\end{tabular}
\label{tab:input_output}
\end{table*}

\begin{table*}[ht]
\small
\centering
\caption{Average length of model responses across rankings for all questions, factoid questions, and non-factoid questions.}
\resizebox{\textwidth}{!}{
    \begin{tabular}{l|llllll}
    \toprule
    \textbf{Method} & rank1 & rank2 & rank3 & rank4 & rank5 & rank6 \\\midrule
   MinosEval predicted (all) & 513.81 & 480.86 & 481.38 & 474.55 & 423.06 & 405.85 \\
    Manually annotated (all) & 573.95 & 498.20 & 477.92	& 446.43 & 408.57 & 374.45\\
   MinosEval predicted (factoid) & 368.64 & 383.83 & 383.91 & 431.11 & 321.36 & 306.58\\
    Manually annotated (factoid)  & 491.84 & 407.29 & 351.40	& 368.35 & 310.30 & 266.24
\\
    MinosEval predicted (non-factoid) & 626.85 & 556.41 & 557.29 & 508.38 & 502.24 & 483.14\\
    Manually annotated (non-factoid) & 636.89 & 568.99 & 576.43 & 507.23 & 485.08 & 458.69 \\
    \bottomrule
    \end{tabular}
}
\label{tab:length-bias}
\end{table*}

\begin{table*}[ht]
\centering
\small
\caption{The samples revisited and reclassified from non-factoid to factoid questions.}
\resizebox{\textwidth}{!}{
    \begin{tabular}{c|p{0.8\textwidth}}  
    \toprule
    \textbf{Dataset} & Sample (question and reference answer) \\\midrule
    \multirow{10}{*}{TREC-DL-NF\_S5} & Question: define visceral? \newline
    Answer: Definition of visceral for English Language Learners: 1. Coming from strong emotions and not from logic or reason. 2. Medical: of or relating to the viscera. \\ \\
    & Question: what is an aml surveillance analyst? \newline
    Answer: The BSA / AML Analyst is responsible for monitoring and investigating customer transactions under applicable anti-money laundering and anti-terrorist financing laws. \\ \\
    & Question: ia suffix meaning? \newline
    Answer: -ia, suffix meaning a specified condition of a disease or process: athrombia, phrenoblabia, pontobulbia. \\\midrule  
    \multirow{10}{*}{ANTIQUE\_S5} & Question: How do I determine the charge of the iron ion in FeCl3? \newline
    Answer: charge of Fe in Fecl3 is 3. Iron has either 2 as valency or 3. In this case, it bonds with three chlorine molecules. Therefore, its valency and charge is three. \\ \\ 
    & Question: What does "see Leaflet" mean on Ept Pregnancy test? \newline
    Answer: It just simply means read the directions that are enclosed with the test if you have any questions. \\ \\
    & Question: What is the difference in a dried plum and a dried prune? \newline
    Answer: A dried prune? That's like saying wet water. Prunes are already dried. A prune is a dried plum, that's all. \\
    \bottomrule
    \end{tabular}
}
\label{tab:trec_antique}
\end{table*}

\newpage

\end{document}